\documentclass{article}
\pdfoutput=1
\usepackage{graphicx}

\PassOptionsToPackage{numbers, compress}{natbib}

\usepackage[final]{neurips_2021}

\usepackage[utf8]{inputenc} 
\usepackage[T1]{fontenc}    
\usepackage{hyperref}       
\usepackage{url}            
\usepackage{booktabs}       
\usepackage{amsfonts}       
\usepackage{nicefrac}       
\usepackage{microtype}      
\usepackage{xcolor}         
\usepackage{algorithm}
\usepackage{algorithmic}
\usepackage{amssymb}
\usepackage{amsmath}

\usepackage{graphicx}
\usepackage{array}
\usepackage{subcaption}

\definecolor{alizarin}{rgb}{0.82, 0.1, 0.26}
\definecolor{auburn}{rgb}{0.43, 0.21, 0.1}
\definecolor{burntorange}{rgb}{0.8, 0.33, 0.0}
\definecolor{cadmiumgreen}{rgb}{0.0, 0.42, 0.24}
\definecolor{cadmiumorange}{rgb}{0.93, 0.53, 0.18}
\definecolor{cadmiumred}{rgb}{0.89, 0.0, 0.13}
\definecolor{cornflowerblue}{rgb}{0.09, 0.58, 0.93}

\newcommand{\eg}{\textit{e.g.},~}
\newcommand{\ie}{\textit{i.e.},~}

\newcommand{\pos}[2]{{$p_{#2}^{#1}$}}
\newcommand{\vel}[2]{{$v_{#2}^{#1}$}}
\newcommand{\ort}[1]{{$\theta^{#1}$}}
\newcommand{\rel}[1]{{$c^{#1}$}}
\newcommand{\bel}[2]{{$h_b^{#1}(#2)$}}
\newcommand{\abo}[2]{{$h_a^{#1}(#2)$}}

\title{Evolution Gym: A Large-Scale Benchmark for Evolving Soft Robots}

\author{%
    Jagdeep Singh Bhatia \\
    MIT CSAIL\\
  \texttt{jagdeep@mit.edu} \\
  \And
    Holly Jackson \\
    MIT CSAIL\\
  \texttt{hjackson@mit.edu} \\
  \And
    Yunsheng Tian \\
    MIT CSAIL\\
  \texttt{yunsheng@csail.mit.edu} \\
  \And
    Jie Xu \\
    MIT CSAIL\\
  \texttt{jiex@csail.mit.edu} \\
  \And
    Wojciech Matusik \\
    MIT CSAIL\\
  \texttt{wojciech@csail.mit.edu} \\
}

\begin{document}

\maketitle

\begin{abstract}

Both the design and control of a robot play equally important roles in its task performance. However, while optimal control is well studied in the machine learning and robotics community, less attention is placed on finding the optimal robot design. This is mainly because co-optimizing design and control in robotics is characterized as a challenging problem, and more importantly, a comprehensive evaluation benchmark for co-optimization does not exist. In this paper, we propose Evolution Gym, the first large-scale benchmark for co-optimizing the design and control of soft robots. In our benchmark, each robot is composed of different types of voxels (e.g., soft, rigid, actuators), resulting in a modular and expressive robot design space. Our benchmark environments span a wide range of tasks, including locomotion on various types of terrains and manipulation. Furthermore, we develop several robot co-evolution algorithms by combining state-of-the-art design optimization methods and deep reinforcement learning techniques. Evaluating the algorithms on our benchmark platform, we observe robots exhibiting increasingly complex behaviors as evolution progresses, with the best evolved designs solving many of our proposed tasks. Additionally, even though robot designs are evolved autonomously from scratch without prior knowledge, they often grow to resemble existing natural creatures while outperforming hand-designed robots. Nevertheless, all tested algorithms fail to find robots that succeed in our hardest environments. This suggests that more advanced algorithms are required to explore the high-dimensional design space and evolve increasingly intelligent robots -- an area of research in which we hope Evolution Gym will accelerate progress. Our website with code, environments, documentation, and tutorials is available at \url{http://evogym.csail.mit.edu}.

\end{abstract}

\section{Introduction}
\label{sec:intro}



One of the main goals of artificial intelligence is to develop effective approaches for the creation of embodied intelligent systems. Inspired from real organisms, where body structure and brain are two key factors for completing any task in a real environment, a successful intelligent robot typically requires concurrently optimizing its structure design and control mechanism. Such a co-design problem has been a long-standing key challenge in the robotics and machine learning communities. Surprisingly, despite its importance, most previous research works still either only develop complex control algorithms for existing robot structures \cite{akkaya2019solving, andrychowicz2020learning, hwangbo2019learning, schulman2017proximal}, or conduct co-optimization over robot morphology and control for only a few simple tasks (\eg running, jumping) \cite{10.1145/2661735.2661737, hejna2021task, sims1994evolving, NEURIPS2019_438124b4}, especially in the soft body domain. The primary reasons behind the under-exploration of co-design algorithms in sophisticated problems are: (1) the underlying complex bilevel optimization scheme of a co-design algorithm, where the inner control optimization loop leads to a long iteration cycle of the whole optimization process; (2) the lack of a well-established benchmark platform providing the researchers with a suite to evaluate and compare different algorithms.


Digital benchmark environments have proven to be successful at promoting the development of advanced learning techniques via providing a comprehensive evaluation suite to make fair comparisons among different algorithms \cite{brockman2016openai, duan2016benchmarking, tassa2018deepmind}. However, to our best knowledge, all existing benchmark platforms constrain their domains within control optimization problems, and the space of co-optimization environment suites is still rarely explored.

To fill this gap, in this work we propose Evolution Gym, a large-scale benchmark for evolving both the shape structure and controller of soft robots. The body of each robot in Evolution Gym is composed of various types of primitive building blocks (\eg soft voxels, rigid voxels, actuator voxels), and the control of the robot includes action signals applied on the actuator voxels. We choose to use this multi-material voxel-based structure as the representation of robot body since it provides a general and universal representation for various categories of robot designs, and at the same time results in a modular and expressive structure design space. We adopt a mass-spring dynamics system \cite{nealen2006physically} with penalty-based frictional contact as the underlining physics engine. Such a light-weight simulator allows the co-design algorithms to significantly reduce the simulation cost and thus accelerate the develop-evaluate iteration cycle \cite{9196808, hiller2014dynamic, MEDVET2020100573}. The back-end simulator is fully developed in C++ to provide further computing efficiency. Another feature of Evolution Gym is its large variety of tasks categorized by varying difficulty levels, which offer an extensive evaluation benchmark for comparing approaches. The benchmark is currently comprised of more than 30 tasks, spanning locomotion on various types of terrains and manipulation. Moreover, Evolution Gym is easy to use. In order to have user-friendly interfaces, we build a Python wrapper outside the C++ simulator and carefully design our APIs off of the well-received APIs of OpenAI Gym with minimum modifications. Evolution Gym will be released fully open-source under the MIT license.

In addition, we develop several baseline algorithms by integrating state-of-the-art design optimization approaches and reinforcement learning techniques. Specifically, in our baseline algorithms, design optimization methods are served in the outer loop to evolve the physical structures of robots and reinforcement learning algorithms are applied in the inner loop to optimize a controller for a given proposed structure design. We conduct extensive experiments to evaluate all baseline algorithms on Evolution Gym. The experiment results demonstrate that intelligent robot designs can be evolved fully autonomously while outperforming hand-designed robots in easier tasks, which reaffirms the necessity of jointly optimizing for both robot structure and control. 
However, none of the baseline algorithms are capable enough to successfully find robots that complete the task in our hardest environments. Such insufficiency of the existing algorithms suggests the demand for more advanced robot co-design techniques, and we believe our proposed Evolution Gym provides a comprehensive evaluation testbed for robot co-design and unlocks future research in this direction.

In summary, our work has the following key contributions: \textbf{(i)} We propose Evolution Gym, the first large-scale benchmark for soft robot co-design algorithms. \textbf{(ii)} We develop several co-design algorithms by combining state-of-the-art design optimization methods and deep reinforcement learning techniques for control optimization. \textbf{(iii)} The developed algorithms are evaluated and analyzed on our proposed benchmark suite, and the results validate the efficacy of robot co-design while pointing out the failure and limitations of existing algorithms.

\section{Related work}

\textbf{Robot co-design}\,\,
Co-designing the structure (\ie body) and control (\ie brain) of robots is a long-standing key challenge in the robotics community. As the earliest work in this space, Sims \cite{sims1994evolving} represents the structure of a rigid robot as a directed graph and proposes an evolutionary algorithm defined on graphs to optimize the robot design. 
Subsequently, the co-design of rigid robots is formulated as a graph search problem where more efficient search algorithms are applied \cite{ha2019reinforcement, pathak2019learning, wang2019neural, 10.1145/3414685.3417831} to achieve increasingly interesting results. 
However, with the restriction of having rigid components only, these algorithms are unable to produce optimal or even feasible designs for many challenging tasks where a compliant joint or robot component is required to achieve the goal. 


On the contrary, soft components offer much more flexibility to represent arbitrary shapes, making the design of more complex, agile, and high-performing robots possible. Inspired by this, some work has been conducted to co-design robots composed of soft cells. Cheney et al. \cite{10.1145/2661735.2661737, corucci2017evolving}; Van Diepen and Shea \cite{van2019spatial}; Corucci et al. \cite{corucci2016evolving}
propose evolutionary algorithms to co-optimize the structure and control of voxel-based robots. However those algorithms typically parameterize the control as an open-loop periodic sequence of actuation, which prevents robots from learning complex non-periodic tasks such as walking on uneven or varying terrains. Spielberg et al. \cite{NEURIPS2019_438124b4} and Medvet et al. \cite{MEDVET2020100573} jointly optimize the spatial-varying material parameters and the neural network policy for soft robots but leave the shape of the robot fixed. Our proposed benchmark shares a similar expressive structure design space as Cheney et al. \cite{10.1145/2661735.2661737}, but allows the control to be parameterized by a sophisticated neural network feedback policy. To handle such sophisticated joint optimization of the robot structure and high-dimensional neural network control policy, we develop several baseline co-design algorithms by combining state-of-the-art design optimization strategies and reinforcement learning techniques for control optmization.

\textbf{Benchmark environments for robotics learning}\,\,
Present research in robotics learning is largely facilitated by emerging benchmark environments. For instance, OpenAI Gym \cite{brockman2016openai}, DeepMind Control Suite \cite{tassa2018deepmind}, rllab \cite{duan2016benchmarking}, and Gibson \cite{xia2018gibson} have been developed to benchmark RL algorithms for controlling rigid robots. At the same time, PlasticineLab \cite{huang2021plasticinelab} is specifically designed for soft robot learning. However, the existing benchmark environments are all constructed for learning the control only. To enable the possibility of evolving the structure of a robot, the existing co-design work has to either implement their own testing environment \cite{NEURIPS2019_438124b4, 10.1145/2661735.2661737, corucci2017evolving,corucci2016evolving, van2019spatial}, or make substantial changes on the underlying code of the existing control-only environments \cite{schaff2019jointly}. The independent development of testing beds requires non-trivial workload, and as a result, existing co-design works mainly focus on evaluating the robot on a few simple tasks such as walking on a flat terrain \cite{10.1145/2661735.2661737, cheney2014evolved, corucci2017evolving, van2019spatial, NEURIPS2019_438124b4, MEDVET2020100573}, or swimming along a single direction \cite{corucci2018evolving, wang2019neural}. An unintended consequence of such independency is an indirect comparison among different algorithms. Evolution Gym fills this gap by presenting a large variety of tasks with different difficulty levels that span from locomotion to manipulation. The proposed benchmark suite can be effectively used to test the generalizability of the algorithms on different tasks, potentially accelerating research in robot co-design.



\section{Evolution Gym}

\begin{figure}[h]
  \vspace{-0.5em}
  \centering
  \includegraphics[width = 0.95\linewidth]{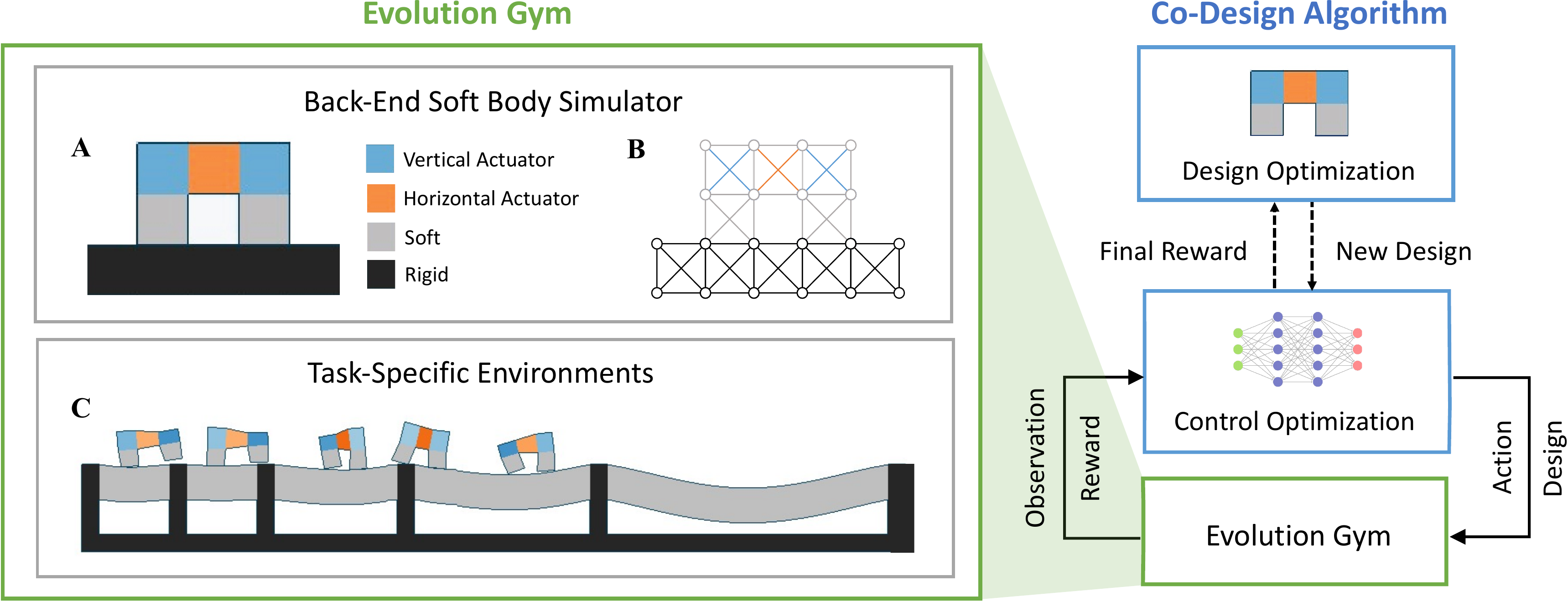}
  \caption{\textbf{Overview of Evolution Gym and its integration with the co-design algorithms.} Evolution Gym is comprised of a back-end soft body simulator (A, B) and task-specific environments (C). A user-customized co-design algorithm can be plugged in to optimize for both robot structure and control through interacting with Evolution Gym on a certain task.
  }
  \label{fig:overview}
  \vspace{-1em}
\end{figure}


\subsection{Overview}
In this section, we present Evolution Gym, a large-scale benchmark for the co-design of voxel-based soft robots. Evolution Gym is featured by its versatile and expressive multi-material voxel-based structure design space, flexibility of the controller parameterization, wide spectrum of tasks of various difficulty levels, fast back-end soft-body simulation support, and user-friendly Python interfaces. 

As shown in the overview in Figure \ref{fig:overview}, Evolution Gym is comprised of a task-specific environment and a back-end soft-body simulator. 
The gym suite provides seamless interfaces with a user-defined co-design algorithm. The co-design algorithm typically consists of a design optimizer and a control optimizer. The design optimizer can propose a new robot structure to the control optimizer, then the control optimizer will compute an optimized controller for the given structure through interactions with Evolution Gym and finally return the maximum reward that this robot structure can achieve. 
In this way, Evolution Gym provides an easy-to-use platform for co-design algorithms to evolve both robot structure and control to optimize for robots' task performances. Evolution Gym is designed to be the first comprehensive testbed for benchmarking and comparing different co-design algorithms with the hope to facilitate the development of more novel and powerful algorithms in the co-design field.






\subsection{Multi-material voxel-based representation}
\label{sec:structure_design_space}

Evolution Gym employs a unified multi-material voxel-based representation for all the components in the environment (\eg robot, terrain, object) as shown in Figure \ref{fig:overview}A. Specifically, each robot in our gym is composed of rigid voxels, soft voxels, horizontal/vertical actuator voxels, and empty voxels. For terrain and objects, we use the same voxel-based structure but with passive voxel types (\ie soft/rigid voxels). 

We chose a voxel-based representation for three main reasons. First, such a multi-material structure of robots provides a general and universal representation for various categories of robot designs and results in a modular structure design space. Additionally, with just the few voxel types described above, and less than 100 voxels per robot, we are able to construct a wide diversity of morphologies due to the resulting combinatorial robot design space. Even with this simple representation, our designed robots are capable of  performing complex motions and completing difficult tasks. Finally, voxel-based robots can be simulated by a fast mass-spring simulation (see section \ref{sec:sim_engine}) which allows our framework to be efficient enough to train robots in a matter of minutes and provides a computationally tractable benchmark for iterating co-design algorithms.



\subsection{Task representation}
Each task in Evolution Gym contains a robot structure proposed by the co-design algorithm, environment specifications (\eg terrain, object), and a task-related goal (\eg locomotion or manipulation). The tasks interface with the co-design algorithm through a few key elements including \textit{robot structure specification}, \textit{observation}, \textit{action}, and \textit{reward}. We introduce each element in detail below.





\textbf{Robot structure specification}\,\,
As described in Section \ref{sec:structure_design_space}, we construct each robot from primitive building blocks arranged on a grid layout. In code, each robot is specified as a material matrix of voxels $\mathcal{M}$ and a connection link list $\mathcal{C}$. The value of entry $m \in \mathcal{M}$ is a label corresponding to a voxel type from the set \{Empty, Rigid, Soft, Horizontal Actuator, Vertical Actuator\}. The connection link list $\mathcal{C}$ stores a list of connection pairs of adjacent voxels. The co-design algorithm can update the robot structure in the environment through \textit{initialization} function with $\mathcal{M}$ and $\mathcal{C}$ as arguments. 

\textbf{Observation}\,\,
The observation is composed in each step to inform the controller of state information of the robot, terrain information of the environment, and goal-relevant information. More specifically, let $N$ be the total number of voxel corner points of the robot. Then the state information of the robot in our tasks is a $(2N + 3)$-D vector including the relative position of each voxel corner with respect to the center of mass of the robot ($2N$-D), and the velocity and orientation of center of mass ($3$-D). 
To handle complex tasks, specifically those with varying terrain types, an additional observation vector including terrain information is provided. We compile terrain information within a local window of size $2W$ around the robot into a length-$2W$ vector observation that describes the terrain's elevation.
Furthermore, goal-related information is offered to inform the controller of the execution status of the current task. This goal-related observation is task-specific and is defined on each task separately. For instance, in manipulation tasks where the robot interacts with some object $O$, we provide orientation and velocity as well as the position of $O$'s center of mass relative to the robot.


\textbf{Action}\,\,
At each time step, an action vector from the robot's controller is provided to step Evolution Gym's simulator. In Evolution Gym, each component of the action vector is associated with an actuator voxel (either horizontal or vertical) of the robot, and instructs a deformation target of that voxel. Specifically, the action value $u$ is within the range $[0.6, 1.6]$, and corresponds to a gradual expansion/contraction of that actuator to $u$ times its rest length.


\textbf{Reward} \,\,
Each task is equipped with a reward function measuring the performance of the current robot and the control action. The value of the reward is defined step-wise and is fed back to the agent through \textit{step} function. The reward function is highly task-specific and should be defined to precisely characterize the robot's completeness of the task. Please refer to Section \ref{sec:tasks} and Appendix for detailed descriptions of the reward functions on each task.



\subsection{Simulation engine}
\label{sec:sim_engine}

We model the dynamics of the underlying simulator as a 2D mass-spring system \cite{nealen2006physically}. This simple, flexible formulation allows us to efficiently model soft robots with a wide range of capabilities in a wide range of environments.  The simulation engine is written entirely in C++. We create Python bindings of our simulator so it seamlessly interfaces with standard learning frameworks.



The simulation represents objects and their environment as a mass-spring system in a grid-like layout (Figure \ref{fig:overview}B).  Objects and their environments are initialized as a set of non-overlapping, connected voxels.  On initialization, each voxel is a cross-braced square, but may undergo deformation as the simulation progresses.  Each edge acts as an ideal spring obeying Hooke's law, with a spring constant defined by one of five possible material types. We employ symplectic RK-4 integration to step forward the simulation. 

Collision detection is performed using a bounding-box tree structure \cite{ericson2004real}. Penalty-based contact forces and frictional forces are computed proportionally to the depth of penetration of the corresponding voxels in contact, and are applied on the voxel vertices in the normal and tangential directions of the contact respectively. Please refer to Appendix \ref{supp:sim} for more details of simulation.

\vspace{-0.2em}
\subsection{Benchmark environment suite}
\label{sec:tasks}
We have developed over 30 unique tasks with Evolution Gym and select 10 tasks here to illustrate the diversity and comprehensiveness of our benchmark task set. All tasks are organized into two categories -- locomotion and manipulation -- though some tasks are a mix of both. We further classify the tasks into different difficulty levels (\ie easy, medium, hard) based on the performance of the baseline algorithms (see Section \ref{sec:algorithm}) on them. We briefly introduce the selected tasks in this section. For more detailed descriptions and visualizations of the tasks, please refer to our website or Appendix \ref{supp:benchmark}. It is also worth mentioning that our gym is designed to be extendable and the user can easily create new tasks for their needs.


\begin{figure}
    \centering
    \includegraphics[width=\textwidth]{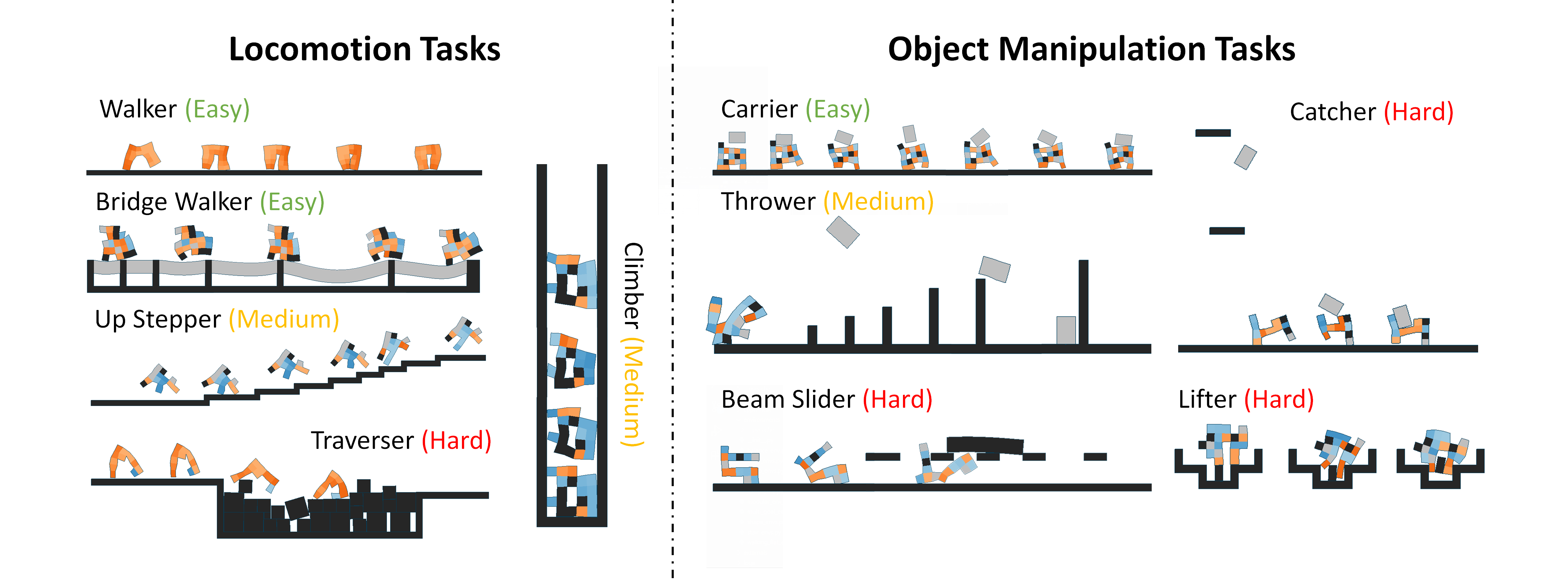}
    \caption{\textbf{A visual overview of selected 10 environments from Evolution Gym.} A verbal description of tasks is provided in Section \ref{sec:tasks}.}
    \label{fig:envs}
    \vspace{-1em}
\end{figure}

\vspace{-0.4em}
\subsubsection{Locomotion tasks}

\textbf{Walker} \textcolor{cadmiumgreen}{(Easy)}  This is a common standard task typically considered by previous works where the robot needs to walk on a flat terrain as fast as possible.

\textbf{Bridge Walker} \textcolor{cadmiumgreen}{(Easy)} In this task, the robot traverses a series of soft “rope” bridges separated by fixed pillars, and similarly as before it needs to maximize its forward speed.

\textbf{Up Stepper} \textcolor{cadmiumorange}{(Medium)} The agent walks up a fixed staircase with steps of varying length.

\textbf{Climber} \textcolor{cadmiumorange}{(Medium)}  The robot must climb two tall fixed walls on each side. The robot is rewarded by its upward climbing speed.

\textbf{Traverser} \textcolor{cadmiumred}{(Hard)}  In this hard task, the robot needs to traverse a pit of rigid blocks to get to the other side without sinking into the pit.

\subsubsection{Object manipulation tasks}

\textbf{Carrier} \textcolor{cadmiumgreen}{(Easy)}  The robot needs to catch a small, soft rectangular object initially dropped from above and then carry it along the forward direction. The robot is rewarded by the distance both it and the object have traveled.

\textbf{Thrower} \textcolor{cadmiumorange}{(Medium)}  The robot throws a soft rectangular box as far as possible without moving itself significantly from its original position.

\textbf{Beam Slider} \textcolor{cadmiumred}{(Hard)} In this task, a beam sits on top of a set of spaced-out floating platforms. The robot is rewarded for moving to the beam and sliding it in the forward direction.

\textbf{Catcher} \textcolor{cadmiumred}{(Hard)} The agent needs to catch a spinning object randomly falling from a high location.

\textbf{Lifter} \textcolor{cadmiumred}{(Hard)} The robot has to manipulate an object and lift it out of a hole.
\section{Evolving soft robots}
\label{sec:algorithm}


Robot evolution/co-design algorithms are formulated as a two-level optimization problem, which involves a design optimization method that evolves physical structures of the robots in the outer loop and a control optimization algorithm that computes an optimized controller for a given robot structure in the inner loop, as illustrated in Algorithm \ref{alg:main}. 
We briefly introduce several instantiations of design optimization methods and control optimization methods in Section \ref{sec:design_optim} and \ref{sec:control_optim} that we use for evaluation on our benchmark, and more details can be found in Appendix \ref{supp:optim}.

\begin{algorithm}
\caption{Algorithmic framework of robot evolution}
    \label{alg:main}
    \begin{algorithmic}
        \STATE \textbf{Inputs:} Task specification $T$, number of generations $n$, population size $p$.
        \STATE \textbf{Outputs:} The best robot design $D^*$ and controller $C^*$.
        \STATE $S \leftarrow \emptyset$ \hfill // Dataset of robot designs, controllers and reward
        \STATE $D_1,...,D_p$ $\leftarrow$ \textsc{SampleDesigns}$(p)$ \hfill // Sample an initial population of robot designs
        \STATE \textbf{for} $i \leftarrow 1$ \TO $n$ \textbf{do}
            \STATE \hskip1.0em \textbf{for} $j \leftarrow 1$ \TO $p$ \textbf{do}
                \STATE \hskip2.0em $C_j \leftarrow$ \textsc{OptimizeControl}$(T, D_j)$ \hfill // Optimize the controller of given robot design
                \STATE \hskip2.0em $r_j \leftarrow$ \textsc{EvaluateReward}$(T, D_j, C_j)$ \hfill // Evaluate the reward of given design and controller
                \STATE \hskip2.0em $S \leftarrow S \cup \{(D_j, C_j, r_j)\}$ \hfill // Update the evaluation result to the dataset
            \STATE \hskip1.0em $D_1,...,D_p$ $\leftarrow$ \textsc{OptimizeDesigns}$(S, p)$ \hfill // Optimize a population of robot designs to evaluate
        \STATE Find the best design $D^*$ and controller $C^*$ in dataset $S$ with the maximum reward $r^*$.
    \end{algorithmic}
\end{algorithm}
\vspace{-0.7em}
\subsection{Design optimization}
\label{sec:design_optim}

Design optimization aims at evolving robot structures to maximize the reward under two physical constraints: the body has to be connected, and actuators must exist. In this section, we introduce three instantiations of the design optimization algorithm (\textsc{OptimizeDesign} in Algorithm \ref{alg:main}). 

\textbf{Genetic algorithm (GA)}\,\,
GAs \cite{michalewicz2013genetic} are widely used in optimizing black-box functions by relying on biologically inspired operators such as mutation, crossover and selection, as demonstrated in previous works on evolving rigid robots \cite{sims1994evolving, wang2019neural}. We implement a simple GA using elitism selection and a simple mutation strategy to evolve the population of robot designs. Specifically, in each generation, our elitism selection works by keeping the top $x\%$ of the robots from the current population as survivors and discarding the rest, where $x$ decreases gradually from 60 to 0 over generations. Next, we iteratively sample and mutate one of those survivors with $10\%$ probability of changing each voxel of the robot to create more offsprings. Note that by mutating a voxel type from/to empty voxel, we are able to change the topology of the robot. The crossover operator is not implemented in our genetic algorithm.

\textbf{Bayesian optimization (BO)}\,\, 
BO \cite{kushner1964new, 10.1007/3-540-07165-2_55} is a commonly used global optimization method for black-box functions by learning and utilizing a surrogate model, which is usually employed to optimize expensive-to-evaluate functions, including evolving rigid robots in previous works \cite{schaff2019jointly, liao2019data}. Specifically, we choose a batch BO algorithm as described in Kandasamy et al. \cite{kandasamy2018parallelised} and implemented in the GPyOpt package \cite{gpyopt2016} that supports categorical input data. We use Gaussian processes as the surrogate model, batch Thompson sampling for extracting the acquisition function, and L-BFGS algorithm to optimize the acquisition function. To ensure a fair comparison with other population-based evolutionary baseline algorithms, the batch size of this algorithm is set equal to the population size of other algorithms.

\textbf{CPPN-NEAT}\,\,
CPPN-NEAT is the predominant method for evolving soft robot design in previous literature \cite{cheney2014evolved, 10.1145/2661735.2661737, corucci2017evolving}. In this method, the robot design is parameterized by a Compositional Pattern Producing Network (CPPN) \cite{stanley2007compositional}. The input to a CPPN is the spatial coordinate of a robot voxel and the output is the type of that voxel. Therefore, by querying the CPPN at all the spatial locations of a robot, we can obtain the type for each voxel to construct a robot. At the same time the NeuroEvolution of Augmenting Topologies (NEAT) algorithm \cite{stanley2002evolving} is used to evolve the structure of CPPNs by working as a genetic algorithm with specific mutation, crossover, and selection operators defined on network structures. Our implementation of CPPN-NEAT is based on the PyTorch-NEAT library \cite{pytorch-neat} and the neat-python library \cite{neat-python}.

\subsection{Control optimization}
\label{sec:control_optim}

In this section, we introduce the specific control optimization algorithm (\textsc{OptimizeControl} in Algorithm \ref{alg:main}) that we use in the robot evolution algorithms. In previous works on evolving soft robots, the controller is either encoded as a fixed periodic sequence of actuation \cite{10.1145/2661735.2661737} or is parameterized as a CPPN that outputs the frequency and phase offset of the periodic actuation for each voxel \cite{corucci2017evolving}. However, the periodic pattern of the control prevents robots from learning complex non-periodic tasks such as walking on uneven or varying terrains. Therefore, we use reinforcement learning (RL) \cite{sutton2018reinforcement} to train the controller, making it possible for the soft robots to perform arbitrarily complex tasks in our benchmark. Specifically, we apply a state-of-the-art RL algorithm named Proximal Policy Optimization (PPO) \cite{schulman2017proximal} for control optimization of robots, with code implementation given by \cite{pytorchrl}. 

\section{Experiments and results}
\label{sec:experiment}

In this section we present the evaluation results of baseline robot co-design algorithms on 10 selected benchmark tasks described in Section \ref{sec:tasks}.
The complete evaluation results on all our benchmark tasks can be found in Appendix \ref{supp:result}.

We develop three baseline algorithms for robot evolution by combing the three design optimization methods in Section \ref{sec:design_optim} and PPO for control optimization in Section \ref{sec:control_optim}. Since the control optimization method is the same for all baseline algorithms, we simply use \textbf{GA}, \textbf{BO}, \textbf{CPPN-NEAT} to denote these three baseline algorithms with different design optimization methods.
The evaluations of our baseline algorithms are performed on machines with Intel Xeon CPU @ 2.80GHz * 80 processors on Google Cloud Platform; GPU is not required. Evaluating one algorithm on a single task usually takes several hours to twenty hours, depending on the number of evaluations, size of population, etc. See Appendix \ref{supp:hyperparam} for more details on hyperparameters of all the experiments.




\subsection{Comparisons among baseline algorithms}

\begin{figure}[h!]
  \centering
  \includegraphics[width=\textwidth]{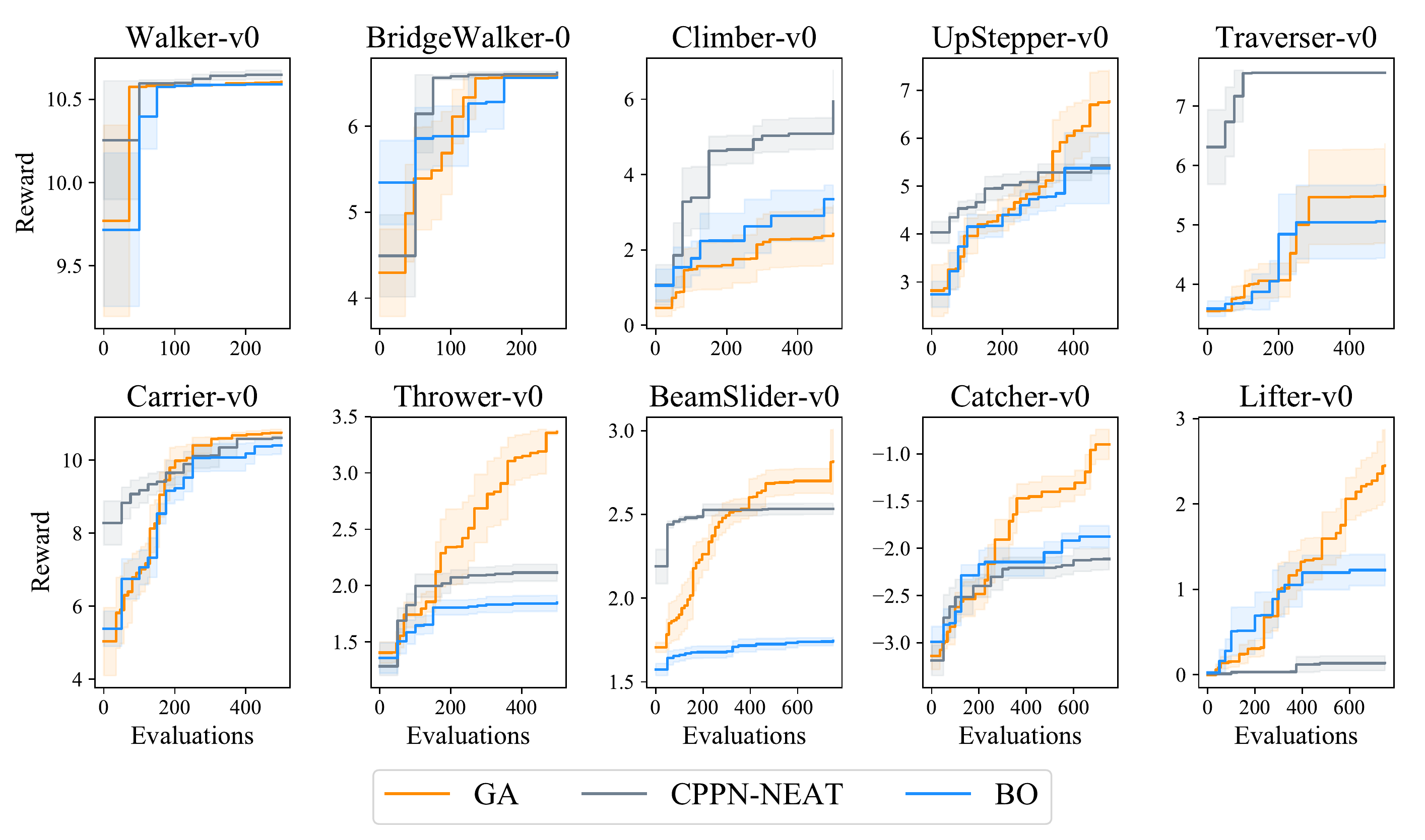}
  \caption{\textbf{Performance comparison among baseline algorithms.} We plot the best performance of robots that each algorithm has evolved w.r.t. the number of evaluations on each task. All the curves are averaged over 6 different random seeds, and the variance is shown as a shaded region.}
  \label{fig:comparison_curves}
\end{figure}

We plot the reward curves of the three baseline algorithms on 10 selected benchmark tasks in Figure \ref{fig:comparison_curves}. There is no single optimal algorithm that performs the best on all tasks, but overall, GA outperforms the other two baseline algorithms. This is surprising because our genetic algorithm is implemented with simple and intuitive operators for mutation and selection without sophisticated mechanisms. Therefore, we believe that with more carefully designed operators, GA has the potential to evolve much more intelligent robots. 
CPPN-NEAT generally performs well on locomotion tasks, as tested by previous works, but performs poorly on more complex manipulation tasks. This is possibly because NEAT favors CPPNs with simpler structures, which encourages CPPNs to generate robots with more regular patterns. However, to succeed in complex manipulation tasks, some agile substructures of the robot must evolve, which might only exist in robots with irregular patterns.
Finally, it is not surprising that BO performs poorly on most of the tasks because the high-dimensional categorical input parameter space and the noisy evaluation done by RL together pose a challenge to fitting an accurate surrogate model in BO.


\begin{figure}[h!]
\vspace{-0.2em}
  \centering
  \includegraphics[width=0.95\textwidth]{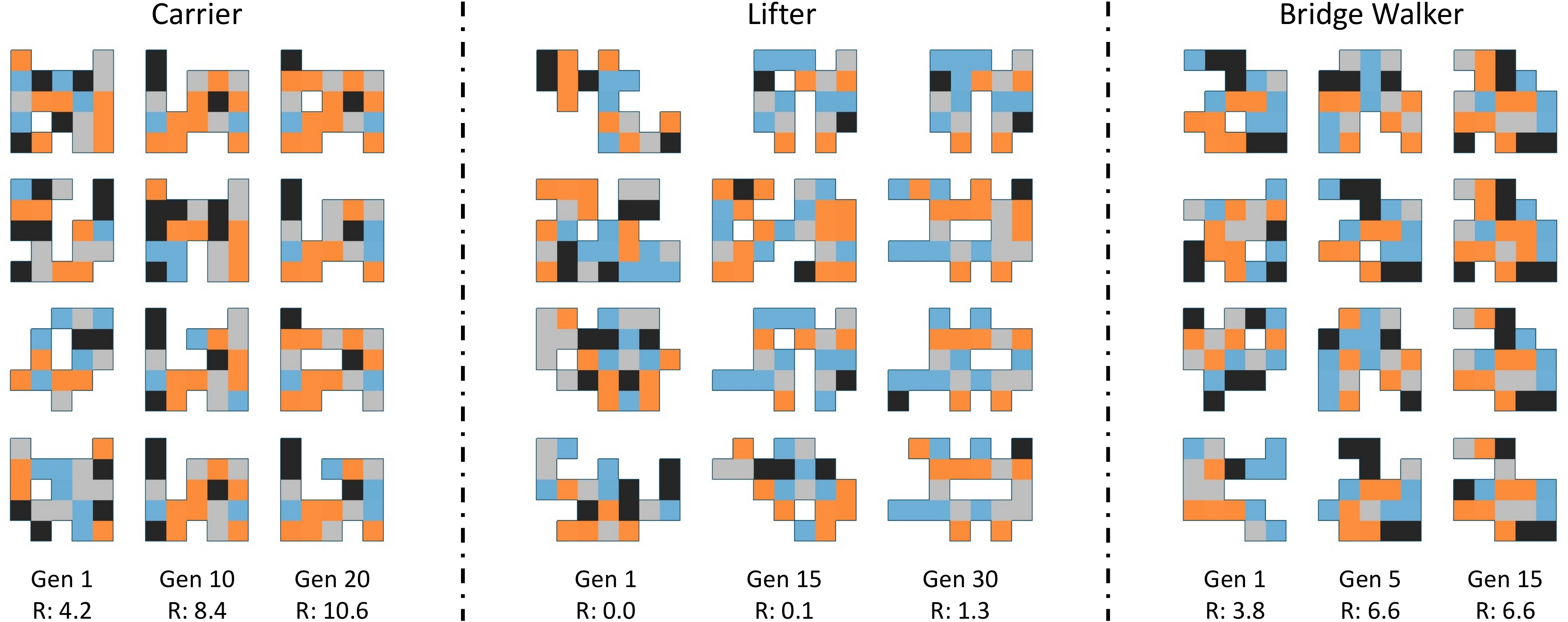}
  \caption{\textbf{Evolution of robot designs.} For each of the three selected tasks, we visualize the population in three different generations. Each column corresponds to one generation for which we show the four top performing robots along with their average reward.}
  \label{fig:evolution_population}
  \vspace{-1em}
\end{figure}

\subsection{Evolution analysis}
In Figure \ref{fig:evolution_population} we visualize the top four robots in three different generations on training the genetic algorithm for the Carrier, Lifter, and Bridge Walker task. We also show the average reward these designs achieve. 

In the carrier task, the robot must catch an object that falls from above and then carry that object as far as possible. Therefore, a successful design for this task achieves two main goals 1) allowing the robot to catch and hold the object securely 2) allowing the robot to move fast. We observe that robots with a block-holding mechanism and with legs are selected for in the top survivors of generation $1$ (randomly initialized). As evolution progresses, these structures become increasingly optimized. Specifically, in later generations, the robots' structures allow them to walk faster while still preventing the block from falling.


A similar comparison pattern can be seen in the Lifter task, where the algorithm learns a parallel gripper-like shape underneath the robot in order to manipulate an object. Unlike in the carrier task, the design structures that the algorithm generates are not prominently found in the initial generation. 
Finally, these patterns are echoed in the Bridge Walker task. Here the robot learns to evolve a large front foot to maximize its surface area and friction force to best walk across the soft rope bridge.  

\subsection{Comparison against hand-designed robots}
\begin{figure}[t!]
  \vspace{-1.5em}
  \centering
  \includegraphics[width=\textwidth]{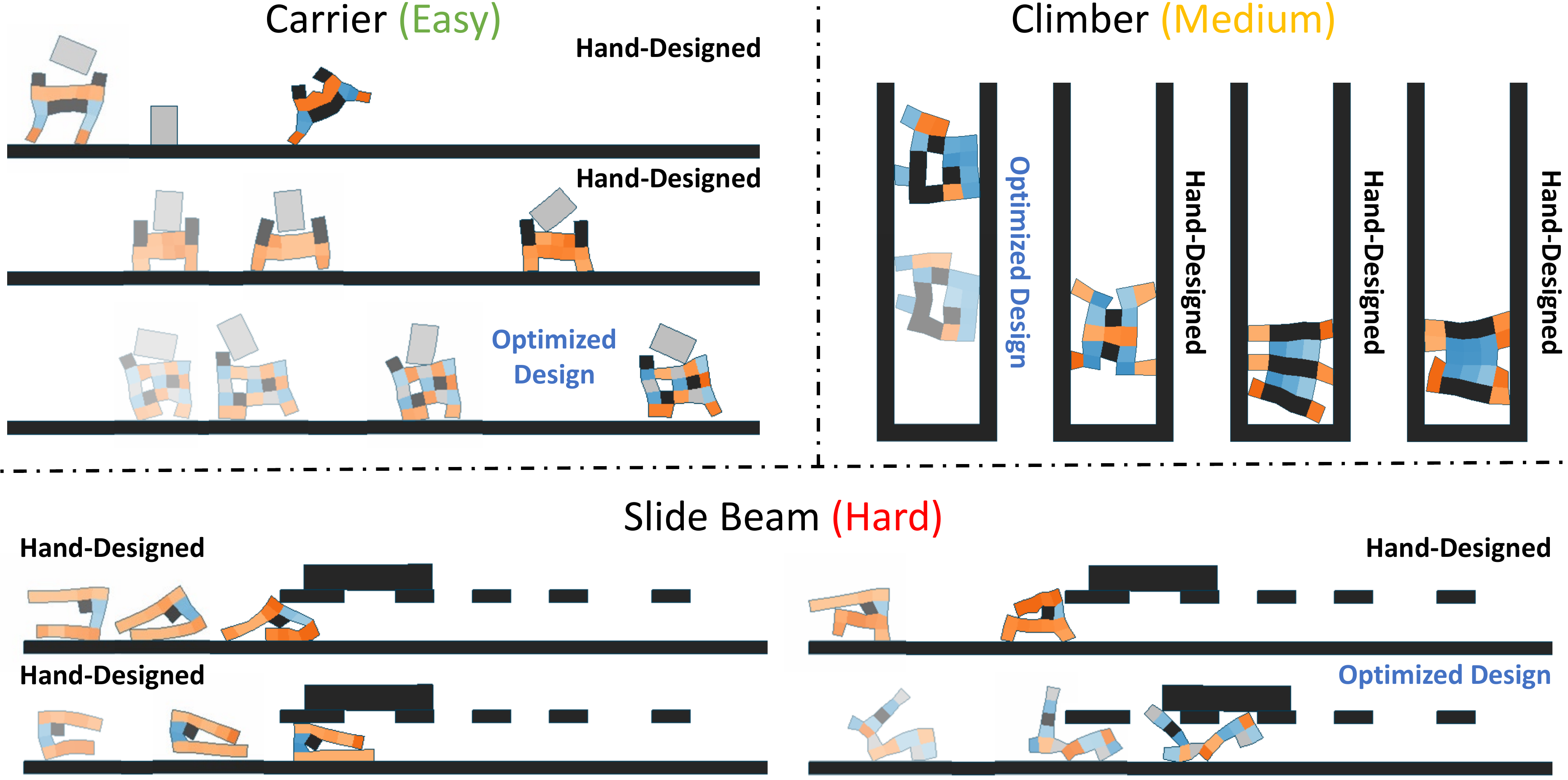}
  \caption{\textbf{Comparison between algorithm-optimized robots and hand designed robots on three tasks}. In each task, we visualize one robot optimized by the algorithm and several hand-designed robots.}
  \label{fig:hand_design_comparison}
  \vspace{-1em}
\end{figure}
We compare the performances of robots optimized by algorithm and the hand designed robots on several tasks to show the necessity of a co-design algorithm (Figure \ref{fig:hand_design_comparison}). The structure of the hand designed robots are bio-inspired and manually constructed according to our best intuition, and their control are optimized by PPO.

For every task, the hand designed robots are outperformed by at least one algorithm (usually more). For instance, for the Climber task we tested numerous natural robot designs. However, none of them successfully climbed very far. The issue with our designs is that we could not find the right trade off between getting traction on the wall, and accelerating upwards. The genetic algorithm, however, is able to find this balance. It develops leg-like structures that help the robot make forward progress, as well as a long flat back that maximizes contact/frictional forces with the wall. Additionally, the genetic algorithm selects for having a hole in the center of its body, which helps it achieve a certain optimized walking motion. 

For other tasks, the performance between the hand designed robots and the robots produced by the algorithms is quite comparable. This is the case with the Carrier robots, as a very natural hand-designed Carrier robot performs almost as well as the best optimized robots produced by the design-optimization algorithms.

In the final case, there are tasks where neither a hand designed nor robot produced by the algorithm could achieve satisfying performance. One such environment is the Beam Slider environment. For this task, many of the hand design robots fail to even achieve the first part of the goal and position themselves underneath the beam. 
While there is one robot produced by the genetic algorithm that does slide the beam across several pegs, from visual observation we believe it comes nowhere close to exhibiting the optimal behavior in this environment. This suggests that further work is needed in designing co-optimization algorithms that can complete these hard tasks.


\section{Conclusion and future work}
\label{sec:conclusion}

In this paper we proposed Evolution Gym, the first large-scale benchmark for evolving the structure and control of soft robots. Through the wide spectrum of tasks in Evolution Gym, we systematically studied the performance of current state-of-the-art co-design algorithms. As a result, we observed how intelligent robots could be evolved autonomously from scratch yet still be capable of accomplishing some surprisingly complex tasks. We also discovered the limitations of existing techniques for evolving more intelligent embodied systems.

There are several potential directions to be explored in the future. First, with the help of our proposed benchmark, it is desirable to develop more advanced co-design algorithms to solve the difficult tasks which existing methods cannot address. Our currently implemented baseline algorithms share a bi-level optimization routine where the design optimization is in the outer loop while the control optimization is in the inner loop. However, Evolution Gym is agnostic to the specific training procedure used. As a result, some ideas for future work using our framework could include concurrently co-optimizing the design and control, neuroevolution algorithms, morphogenetic development, gradient-based methods for design optimization, or algorithms with decentralized controllers.

Second, a robot will be considered more successful if it can perform multiple tasks. Our benchmark suite naturally provides a comprehensive set of tasks and can potentially promote more exciting research work about multi-task or multi-objective robot co-design algorithms. 

Another consideration is the specific morphological encodings used by the codesign algorithms as more intelligent encodings could lead to better performance. For instance, \cite{veenstra2020different} analyzes the strengths and weaknesses of different morphological encodings. Our baseline algorithms use a direct encoding and CPPN but exploring other encoding representations remains interesting future work. 

Finally, since tasks in Evolution Gym are currently limited to either locomotion or manipulation, we plan to further extend Evolution Gym to additional task categories such as flying or swimming by incorporating new simulation capabilities.

Overall, we believe our carefully-designed benchmarking tool fills an important missing piece in research in soft robotics and robotic evolution algorithms. Armed with the flexible and expressive framework Evolution Gym provides, we are optimistic that future researchers will use Evolution Gym as a standard test bed to improve co-design methods and evolve more intelligent robots.

\section*{Societal Impact}
\label{sec:societal}

We regard this work as a very preliminary piece of research in the field of soft robot co-design, and therefore think that we are still far away from causing harm to society. However, we can definitely foresee some problems if this technology were to be applied in the real world on a large scale. For instance, this work may inspire the automatic design of real biological creatures in which serious ethical issues exist. Additionally, since the users have full control over the reward design when customizing the benchmark environments, they could specify pernicious goals and encourage the co-design algorithm to produce more biased results.

\begin{ack}
We thank Tao Du and the anonymous reviewers for their helpful comments in revising the paper. This work is supported by the Defense Advanced Research Projects Agency (FA8750-20-C-0075). 
\end{ack}


\bibliographystyle{plain}
\bibliography{bibliography}

\newpage

\appendix

\section{Simulation engine}
\label{supp:sim}

In this section we describe Evolution Gym's simulator in detail. We describe the simulator's representation of objects, the dynamics of the underlying simulator, our implementation of contact forces, and other techniques we use to improve the quality of Evolution Gym. Finally, we present an analysis of how our simulation scales with the number of voxels, followed by some of the hyperparamters in our implementation.   

\subsection{Representation}

\begin{figure}[h!]
\vspace{-0.2em}
  \centering
  \includegraphics[width=0.95\textwidth]{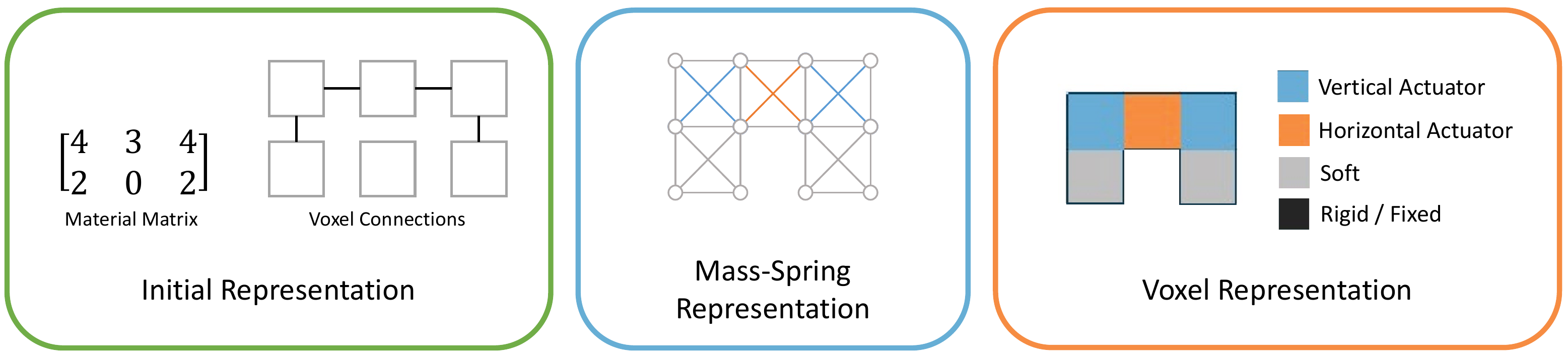}
  \caption{Representation of Simulation Objects}
  \label{fig:rep}
  \vspace{-0em}
\end{figure}


The simulation represents objects and their environment as a 2D mass-spring system in a grid-like layout, where objects are initialized as a set of non-overlapping, connected voxels.

More specifically,  any objects loaded into the simulation, including the robot, can be represented as a material matrix and a set of connections between adjacent cells as seen in the green panel of Figure \ref{fig:rep}. The entries of the material matrix are integers corresponding to a voxel type from the set \{Empty, Rigid (black), Soft \textcolor{lightgray}{(gray)}, Horizontal Actuator \textcolor{cadmiumorange}{(orange)}, Vertical Actuator \textcolor{cornflowerblue}{(blue)}, Fixed (black)\}. 
In particular, actuator cells are specific to the robot object and fixed cells are only used in non-robot objects. Each voxel can be connected to each of its adjacent voxels, which determines the identity of objects in the simulation. In our work, we connect all pairs of adjacent voxels. 

The simulation converts all objects into a set of point masses and springs by turning each voxel into a cross-braced square as shown in the blue panel of Figure \ref{fig:rep}. Note that some voxels share the same point masses and springs. All point masses in the simulation have the same mass and the equilibrium lengths of axis-aligned and diagonal springs are constants for simplicity. However, the spring constants assigned vary based on voxel material-type -- with ties broken in favor of the more-rigid spring. Please see section \ref{sim:hyper} for more details on simulation hyperparameters.

\subsection{Simulation Dynamics}
Let $y_n$ be the state of the simulation at time step $n$. We can view $y_n$ as a length $2 \times 2N$ vector of 2D positions and velocities, where $N$ is the number of point masses in the simulation. We use symplectic RK-4 integration to compute $y_{n+1}$ by taking into account gravitational, contact, viscous drag, and spring forces. 





In particular, spring forces are modelled by the following dynamics equations:
\begin{align*}
    &\mathbf{f}_{int} + \mathbf{f}_{ext} = M\ddot{\mathbf{x}}\\
    &\mathbf{f}_{int}^i = \sum_jk_j(l_j - \bar{l}_j)\mathbf{e}^i_j, \,\,\,j \in \text{springs associated with vertex }i
\end{align*}
where $\mathbf{x}$ is the positions of all vertices in the mass-spring system, $M$ is the mass matrix of system, $\mathbf{f}_{int}$ is the internal spring forces exerted on vertices, $\mathbf{f}_{ext}$ is the external forces produced by gravity and contacts, $k_j$ is the spring constant in Hooke's law, $l_j$ and $\bar{l}_j$ are the lengths of spring $j$ in current shape and in rest shape respectively, and $\mathbf{e}^i_j$ is the unit direction of the spring $j$ relative to vertex $i$. 

For our simulator, we use $50$Hz as the control frequency, while using $1500$Hz as the simulation sub-step frequency. This is achieved by running $30$ sub-steps of the simulation for each control input. We apply such sub-step scheme to improve the stability of the simulation.

\subsection{Collision detection and contact forces}
We use bounding box trees for collision detection, made up of the voxels on the surface of each object  \cite{ericson2004real}. Exploiting the grid-like nature of voxels, the tree for each object is only computed at initialization. However, the bounding box at each node of the tree is recomputed at each time step.



To resolve collisions between cells, penalty-based contact forces and frictional forces are computed proportionally to the depth of penetration of the corresponding cells in contact. These forces are applied on voxel vertices in the normal and tangential directions of the contact respectively.

It is important to note that contact forces are pre-computed before each RK-4 step and are considered constant by the time step integrator. We chose to do this because computing contact forces is often computationally expensive.


\subsection{Other techniques}
We use several other techniques to improve the quality of the simulation engine and its interactions with the control optimization.

\textbf{Strain limiting}\\
Strain limiting is implemented to prevent the self-folding of objects in the simulation. If any spring grows or shrinks by more than $25\%$, the simulation will reposition the masses connected by the spring to reduce the strain. The springs of rigid cells have a more aggressive threshold for strain limiting at $3\%$ compression/expansion. Note that strain limiting does not impose a hard limit on springs and can still be overcome by very strong actuations from the robot.

\textbf{Self-folding}\\
Even with strain limiting and collision detection, it is still possible that the robot in the simulation can fold in on itself. In order to combat this, we have a reliable check for whether the robot object is self-folding: we check whether the number of colliding, non-adjacent pairs of voxels on the surface of the robot is more than the number of voxels on the surface of the robot. In the each of our environments described in Section \ref{supp:benchmark}, we penalize the robot with a one time reward of $-3$ and reset the environment any time self-folding is detected.

\textbf{Delayed actuations}\\
Each time the environment steps, the robot's controller provides a single actuation value for each of the robot’s actuators. In order to move the actuators, the simulation changes the equilibrium lengths of the actuator's springs. However, rather than setting the goal equilibrium length right away, the simulation sets the equilibrium length to be the weighted average of the goal and the spring's current length by parameters $\alpha$ and $1-\alpha$, respectively, for $\alpha << 0.5$. This has the benefit of requiring the robot's controller to favor longer smoother motions -- which are more in line with how a real actuator might behave -- rather than short abrupt ones.

\subsection{Scalability and speed analysis}

\begin{figure}[]
    \centering
    \includegraphics[width=1.0\textwidth]{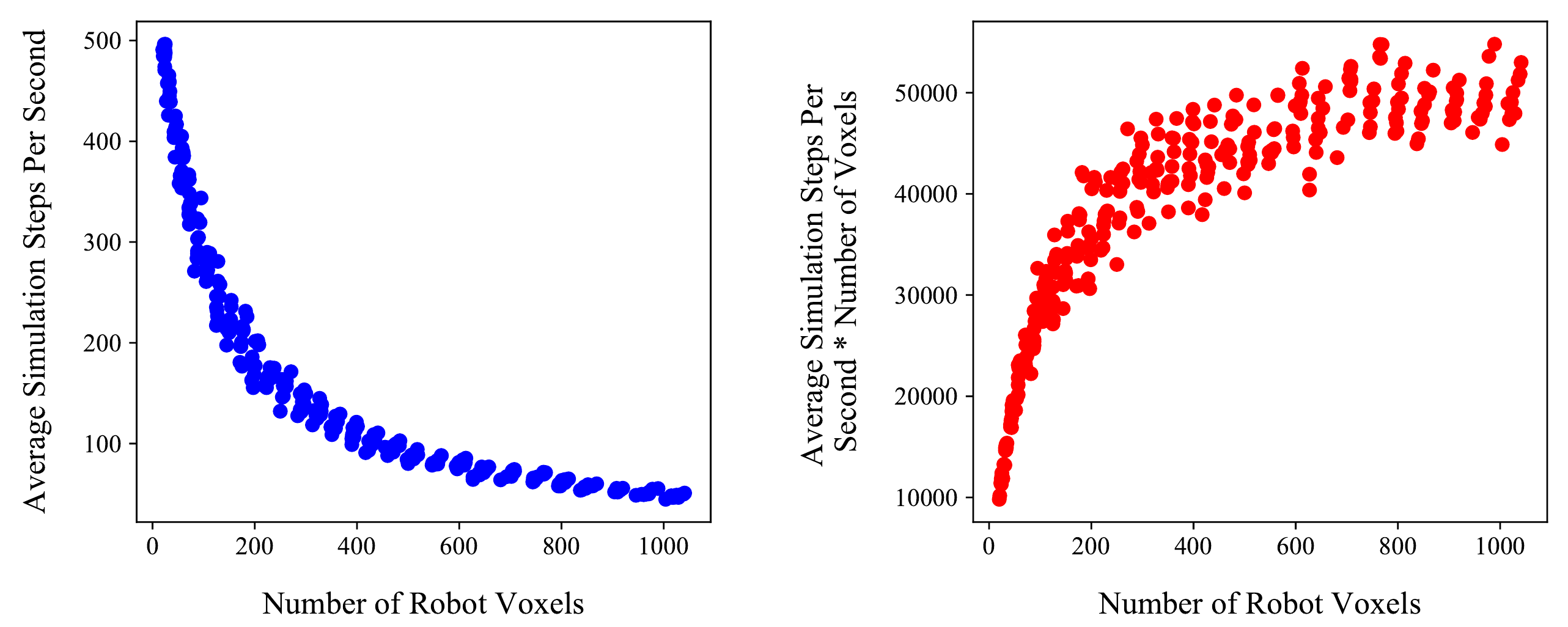}
    \caption{Simulating robots with varying voxel counts on a single core of an Intel Xeon CPU @ 2.80GHz. We graph the average number of simulation steps per second (left) and the average number of simulation steps per second times the number of voxels (right).}
    \label{fig:speed}
\end{figure}

As co-design algorithms become more advanced, we should expect them to search over larger design spaces. Therefore, we believe that the ability of our simulation to scale well as the number of voxels increases is important. In this sub-section we test the performance of our simulator at varying numbers of voxels and present the result in Figure \ref{fig:speed}.

Specifically, our experiment is as follows: For each $n \in [5, 35]$, we sample $10$, $(n \times n)$ robots and evaluate the performance of our simulator when simulating these robots for $10$ seconds and with random actions. Robots are simulated in a simple environment with flat terrain consisting of $100$ voxels. Simulations are run on a single core of an Intel Xeon CPU @ 2.80GHz.

From these experiments, we compare the \hspace{0.4em}\texttt{\# of non-empty voxels} \hspace{0.4em} against \hspace{0.4em}\texttt{average number of simulation steps per second (assps)}. We also compute \hspace{0.4em} \texttt{assps} $\times$ \texttt{\# of non-empty voxels}\hspace{0.4em} as another metric.

These results allow us to compare our simulator to others. For instance, our simulation speed is slower than most rigid-body simulations, like MuJoCo \cite{todorov2012mujoco}. This is because simulation speed is very dependent on the type of simulation, and there are many more degrees of freedom in soft body simulation like ours compared to a standard rigid body simulation.
	
Another insightful comparison might be with soft, 3D FEM-based simulations. For example, in Du et al. \cite{du2021diffpd}, one simulation step consumes on the order of several seconds to minutes and the PPO training for a single robot design takes on the order of hours to converge. By comparison, in our framework, a single robot can be simulated with hundreds of steps per second and trained in parallel on a 4-core machine in a matter of minutes. This highlights the advantage of our 2D mass-spring approach.

Finally, the best comparison of our work would be other 2D voxel-based simulations, of which few exist. For example, our speed is comparable to Medvet et al. \cite{medvet2020evolution} as we both simulate objects in 2D and adopt relatively standard implementations for mass-spring systems. However, the main contribution of Medvet et al. is its simulation; our main advantages over Medvet et al 2020 are (1) a comprehensive set of well-designed tasks with various difficulty levels to provide the first comparison platform for evaluating co-design algorithms, (2) our implementation of state-of-the-art algorithms to establish a baseline for co-design, and (3) our ability to interface with standard python ML libraries through python bindings for our physics simulation.

\subsection{Hyperparameters}
\label{sim:hyper}

\begin{table}[h!]
  \caption{Values of Simulation hyperparameters}
  \label{tab:simhp}
  \centering
  \begin{tabular}{ll}
    \toprule
    \cmidrule(r){1-2}
    parameter name     & value \\
    \midrule
        point mass & $1.0$\\
        viscous drag & $0.1$ \\
        gravity & $110$\\
        \cmidrule(r){1-2}
        contact stiffness coefficient & $2.1 \cdot 10^7$\\
        collision penetration depth additive & $5 \cdot 10^{-3}$\\
        coefficient of friction & $0.1$\\
        frictional penalty factor & $0.5$\\
        friction multiplier & $2.4 \cdot 10^3$\\
        \cmidrule(r){1-2} 
        rigid main spring const & $\frac{3 \cdot 10^8}{3.5}$\\
        rigid structural spring const & $\frac{3 \cdot 10^8}{7}$\\
        soft main spring const & $\frac{3 \cdot 10^8}{5}$\\
        soft structural spring const & $\frac{3 \cdot 10^8}{10}$\\
        actuator main spring const & $\frac{3 \cdot 10^8}{6}$\\
        actuator structural spring const & $\frac{3 \cdot 10^8}{24}$\\
    \bottomrule
  \end{tabular}
\end{table}


In this section we describe the significance of all hyperparameters on the simulation.

We start with some general hyperparameters. The \textit{point mass} constant describes the mass of all points. 
The \textit{viscous drag} constant is a multiplier used to adjust the strength of the viscous drag force. Increasing this constant makes the particles move as though they are traveling through a more viscous fluid. The \textit{gravity} constant describes the magnitude of the force of gravity in the simulation.

We continue with hyperparameters important to contact forces. The \textit{contact stiffness coefficient} is a multiplier used to adjust the strength of the normal contact force between objects. The \textit{collision penetration depth additive} is a constant added to the penetration depth of collision when contact forces are computed. \textit{Coefficient of friction} and \textit{frictional penalty factor} correspond to the coefficients of static and dynamic friction, respectively, while \textit{friction multiplier} is a constant used to adjust the strength of the tangential contact force between objects.

Finally, we describe the hyperparameters which control spring rigidity. Increasing any of these constants makes the corresponding springs more rigid, and decreasing them has the opposite effect. Recall that in the simulation each voxel is a cross-braced square. \textit{Main} spring constants describe the rigidity of springs around the square edges of a voxel while \textit{structural} spring constants describe the rigidity of springs on the cross-brace. \textit{Rigid, soft, and actuator} spring constants correspond to springs on rigid, soft, and actuator cells, respectively. 
\section{Full benchmark suite}
\label{supp:benchmark}
We have implemented a total of $32$ tasks for Evolution Gym. Below, we describe in detail the reward and observation of each of the environments we have implemented. 

For reference, the names of the $10$ benchmark tasks are 
\texttt{Walker-v0}, \texttt{BridgeWalker-v0}, \texttt{UpStepper-v0}, \texttt{Traverser-v0}, \texttt{Climber-v0},  \texttt{Carrier-v0},  \texttt{Thrower-v0},  \texttt{Catcher-v0},  \texttt{Lifter-v0},  \texttt{BeamSlider-v0}.


\subsection{Notation}
We start by describing some notation that we will use in the following sections.

\textbf{Position}\\
Let \pos{o}{} be a vector of length $2$ that represents the position of the center of mass of an object $o$ in the simulation at time $t$. \pos{o}{x} and \pos{o}{y} denote the $x$ and $y$ components of this vector, respectively. \pos{o}{} is computed by averaging the positions of all the point-masses that make up object $o$ at time $t$.

\textbf{Velocity}\\
Similarly, let \vel{o}{} be a vector of length $2$ that represents the velocity of the center of mass of an object $o$ in the simulation at time $t$. \vel{o}{x} and \vel{o}{y} denote the $x$ and $y$ components of this vector, respectively. \vel{o}{} is computed by averaging the velocities of all the point-masses that make up object $o$ at time $t$.

\textbf{Orientation}\\
Similarly, let \ort{o} be a vector of length $1$ that represents the orientation of an object called $o$ in the simulation at time $t$. Let $p_i$ be the position of point mass $i$ of object $o$. We compute \ort{o} by averaging over all $i$ the angle between the vector $p_i - \text{\pos{o}{}}$ at time $t$ and time $0$. This average is a weighted average weighted by $||p_i - \text{\pos{o}{}}||$ at time $0$.

\textbf{Special observations}\\
Let \rel{o} be a vector of length $2n$ that describes the positions of all $n$ point masses of object $o$ relative to \pos{o}{}. We compute \rel{o} by first obtaining the $2 \times n$ matrix of positions of all the point masses of object $o$, subtracting \pos{o}{} from each column, and reshaping as desired.

Let \bel{o}{d} be a vector of length $(2d+1)$ that describes elevation information around the robot below its center of mass. More specifically, for some integer $x \le d$, the corresponding entry in vector \bel{o}{d} will be the highest point of the terrain which is less than \pos{o}{y} between a range of $[x, x+1]$ voxels from \pos{o}{x} in the $x$-direction.

Let \abo{o}{d} be a vector of length $(2d+1)$ that describes elevation information around the robot above its center of mass. More specifically, for some integer $x \le d$, the corresponding entry in vector \abo{o}{d} will be the lowest point of the terrain which is greater than \pos{o}{y} between a range of $[x, x+1]$ voxels from \pos{o}{x} in the $x$-direction.

\subsection{Walking tasks}


\subsubsection{Walker-v0}

\begin{figure}[H]
    \centering
    \includegraphics[width=0.8\textwidth]{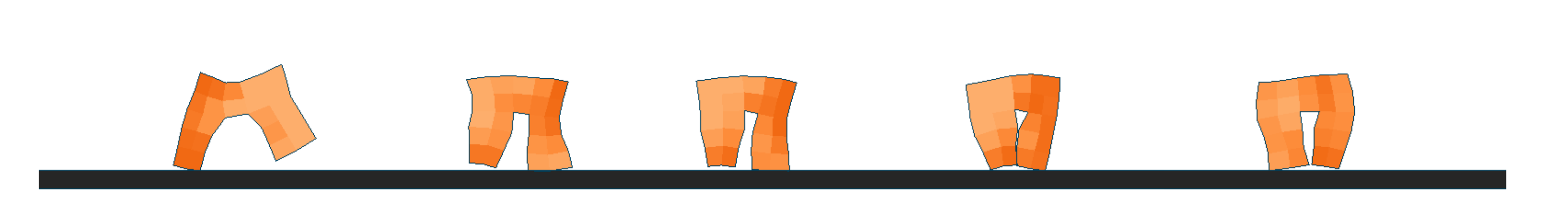}
    \caption{\texttt{Walker-v0}}
    \label{fig:walking}
\end{figure}

In this task the robot walks as far as possible on flat terrain.  This task is \textcolor{cadmiumgreen}{\textbf{easy}}.

Let the robot object be $r$. The observation space has dimension $\mathcal{S} \in R^{n + 2}$, where $n$ is the number of point masses in object $r$, and is formed by concatenating vectors 
$$\text{\vel{r}{}, \rel{r}}$$
with lengths $2$ and $n$, respectively. The reward $R$ is 
$$R = \Delta\text{\pos{r}{x}}$$
which rewards the robot for moving in the positive $x$-direction.

This environment runs for $500$ steps. The robot also receives a one-time reward of $1$ for reaching the end of the terrain.

\subsubsection{BridgeWalker-0}

\begin{figure}[H]
    \centering
    \includegraphics[width=0.8\textwidth]{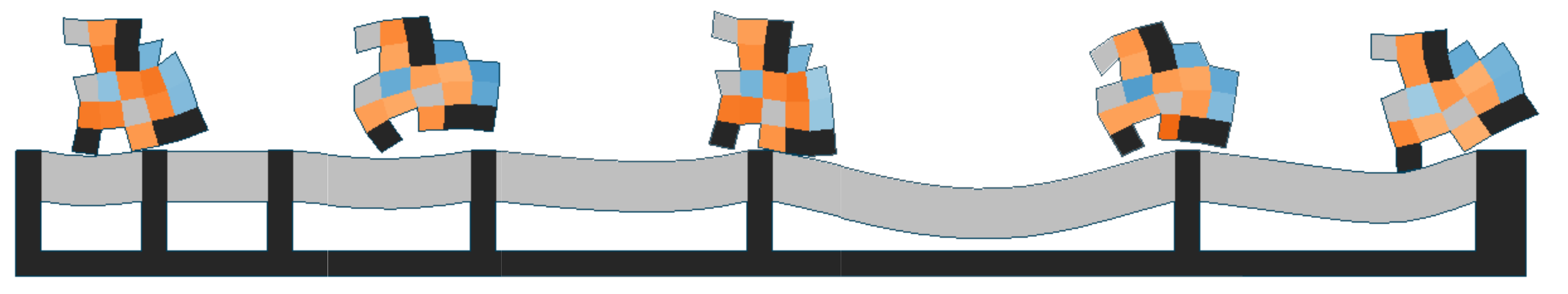}
    \caption{\texttt{BridgeWalker-v0}}
    \label{fig:soft_bridge}
\end{figure}

In this task the robot walks as far as possible on a soft rope-bridge. This task is \textcolor{cadmiumgreen}{\textbf{easy}}.

Let the robot object be $r$. The observation space has dimension $\mathcal{S} \in R^{n + 3}$, where $n$ is the number of point masses in object $r$, and is formed by concatenating vectors 
$$\text{\vel{r}{}, \ort{r}, \rel{r}}$$
with lengths $2$, $1$, and $n$, respectively. The reward $R$ is 
$$R = \Delta\text{\pos{r}{x}}$$
which rewards the robot for moving in the positive $x$-direction. The robot also receives a one-time reward of $1$ for reaching the end of the terrain. 

This environment runs for $500$ steps.

\subsubsection{BidirectionalWalker-v0}

\begin{figure}[H]
    \centering
    \includegraphics[width=0.8\textwidth]{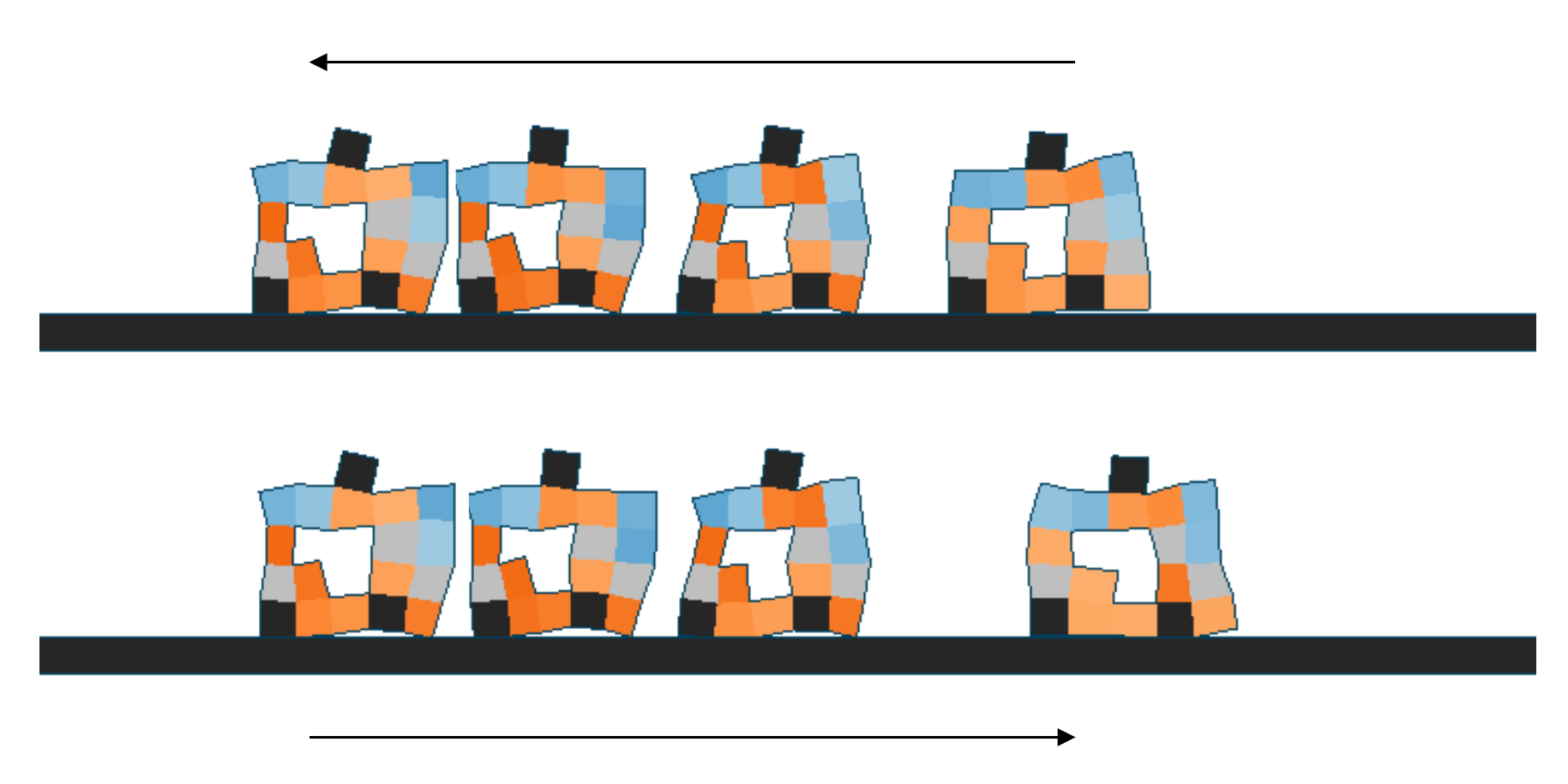}
    \caption{\texttt{BidirectionalWalker-v0}}
    \label{fig:biwalk}
\end{figure}

In this task the robot walks bidirectionally.  This task is \textcolor{cadmiumorange}{\textbf{medium}}.

Let the robot object be $r$. Let $g_x$ be a goal $x$-position that is randomized and changes throughout the task. There is also a counter $c$ which counts how many times the goal has changed. The observation space has dimension $\mathcal{S} \in R^{n + 5}$, where $n$ is the number of point masses in object $r$, and is formed by concatenating vectors 
$$\text{\vel{r}{}},\, \text{\rel{r}},\, c,\, g_x,\, g_x-\text{\pos{r}{x}}$$
with lengths $2$, $n$, $1$, $1$, and $1$, respectively. The reward $R$ is 
$$R = -\Delta|g_x - \text{\pos{r}{x}}|$$
which rewards the robot for moving towards the goal in the $x$-direction.

This environment runs for $1000$ steps.

\subsection{Object manipulation tasks}

\subsubsection{Carrier-v0}

\begin{figure}[H]
    \centering
    \includegraphics[width=0.8\textwidth]{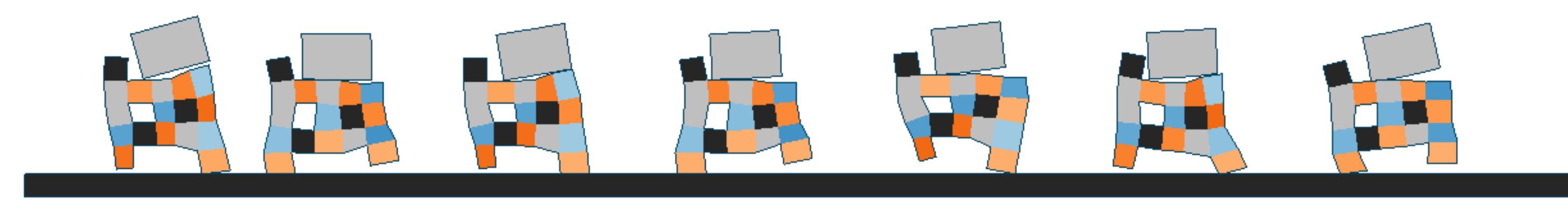}
    \caption{\texttt{Carrier-v0}}
    \label{fig:carry_small_rect}
\end{figure}

In this task the robot catches a box initialized above it and carries it as far as possible. This task is \textcolor{cadmiumgreen}{\textbf{easy}}.

Let the robot object be $r$ and the box object the robot is trying to carry be $b$. The observation space has dimension $\mathcal{S} \in R^{n + 6}$, where $n$ is the number of point masses in object $r$, and is formed by concatenating vectors 
$$\text{\vel{b}{}},\, \text{\pos{b}{}} - \text{\pos{r}{}},\, \text{\vel{r}{}, \, \text{\rel{r}}}$$
with lengths $2$, $n$, $2$, and $2$, respectively. The reward $R = R_1 + R_2$ is the sum of several components. 
$$R_1 = 0.5 \cdot \Delta\text{\pos{r}{x}}  +  0.5 \cdot \Delta\text{\pos{b}{x}}$$
which rewards the robot and box for moving in the positive $x$-direction.
$$R_2 = 
\begin{cases} 
    0 & \mbox{if } \text{\pos{b}{y}} \ge t_y \\
10 \cdot \Delta\text{\pos{b}{y}} & \text{otherwise} 
\end{cases}$$
which penalizes the robot for dropping the box below a threshold height $t_y$.

This environment runs for $500$ steps. The robot also receives a one-time reward of $1$ for reaching the end of the terrain.

\subsubsection{Carrier-v1}

\begin{figure}[H]
    \centering
    \includegraphics[width=0.8\textwidth]{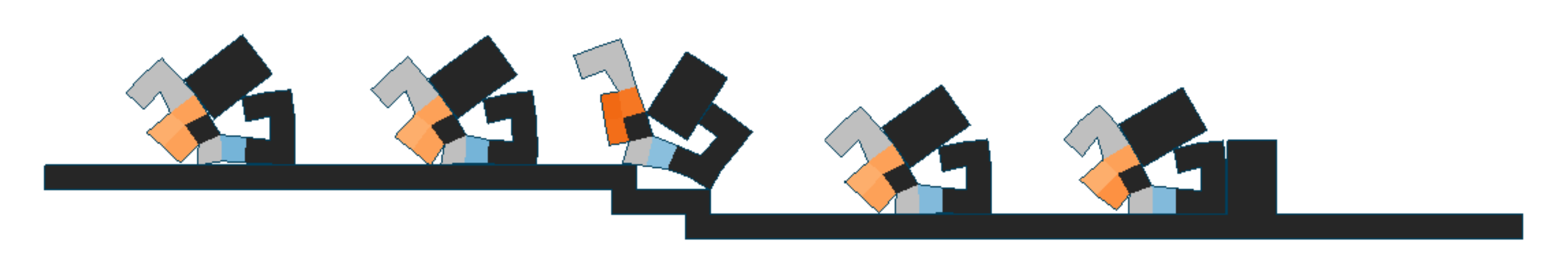}
    \caption{\texttt{Carrier-v1}}
    \label{fig:carry_small_rect1}
\end{figure}

In this task the robot carries a box to a table and places the box on the table. This task is \textcolor{cadmiumred}{\textbf{hard}}.

Let the robot object be $r$ and the box object the robot is trying to carry be $b$.  We achieve the described behavior by setting a goal $x$-position - $g^r_x$ and $g^b_x$ - for the robot and box, respectively. The observation space has dimension $\mathcal{S} \in R^{n + 6}$, where $n$ is the number of point masses in object $r$, and is formed by concatenating vectors 
$$\text{\vel{b}{}},\, \text{\pos{b}{}} - \text{\pos{r}{}},\, \text{\vel{r}{}, \, \text{\rel{r}}}$$
with lengths $2$, $n$, $2$, and $2$, respectively. The reward $R = R_1 + R_2+R_3$ is the sum of several components. 
$$R_1 = -2 \cdot \Delta|g^b_x - \text{\pos{b}{x}}|$$
which rewards the box for moving to its goal in the $x$-direction.
$$R_2 = -\Delta|g^r_x - \text{\pos{r}{x}}|$$
which rewards the box for moving to its goal in the $x$-direction.
$$R_3 = 
\begin{cases} 
    0 & \mbox{if } \text{\pos{b}{y}} \ge t_y \\
10 \cdot \Delta\text{\pos{b}{y}} & \text{otherwise} 
\end{cases}$$
which penalizes the robot for dropping the box below a threshold height $t_y$. Note that in this task $t_y$ is not constant, and varies with the elevation of the terrain.

This environment runs for $1000$ steps.

\subsubsection{Pusher-v0}

\begin{figure}[H]
    \centering
    \includegraphics[width=0.8\textwidth]{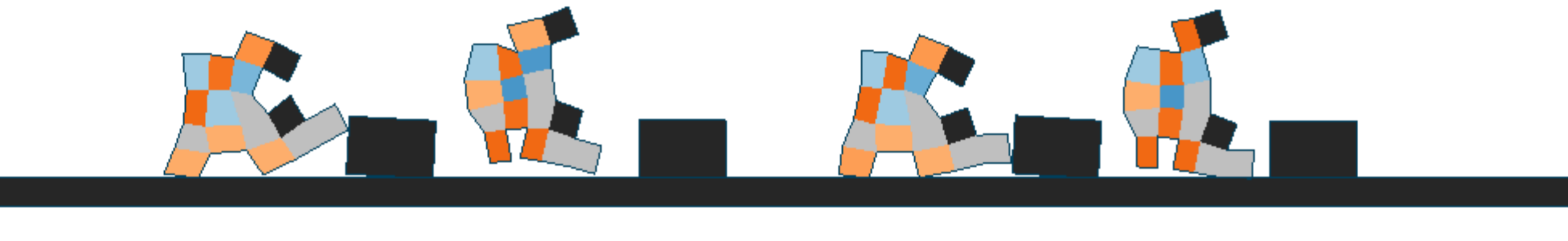}
    \caption{\texttt{Pusher-v0}}
    \label{fig:push_small_rect}
\end{figure}

In this task the robot pushes a box initialized in front of it. This task is \textcolor{cadmiumgreen}{\textbf{easy}}.

Let the robot object be $r$ and the box object the robot is trying to push be $b$. The observation space has dimension $\mathcal{S} \in R^{n + 6}$, where $n$ is the number of point masses in object $r$, and is formed by concatenating vectors 
$$\text{\vel{b}{}},\, \text{\pos{b}{}} - \text{\pos{r}{}},\, \text{\vel{r}{}, \, \text{\rel{r}}}$$
with lengths $2$, $n$, $2$, and $2$, respectively. The reward $R = R_1 + R_2$ is the sum of several components. 
$$R_1 = 0.5 \cdot \Delta\text{\pos{r}{x}}  +  0.75 \cdot \Delta\text{\pos{b}{x}}$$
which rewards the robot and box for moving in the positive $x$-direction.
$$R_2 = -\Delta|\text{\pos{b}{x}} - \text{\pos{r}{x}}|$$
which penalizes the robot and box for separating in the $x$-direction.

This environment runs for $500$ steps. The robot also receives a one-time reward of $1$ for reaching the end of the terrain.

\subsubsection{Pusher-v1}

\begin{figure}[H]
    \centering
    \includegraphics[width=0.8\textwidth]{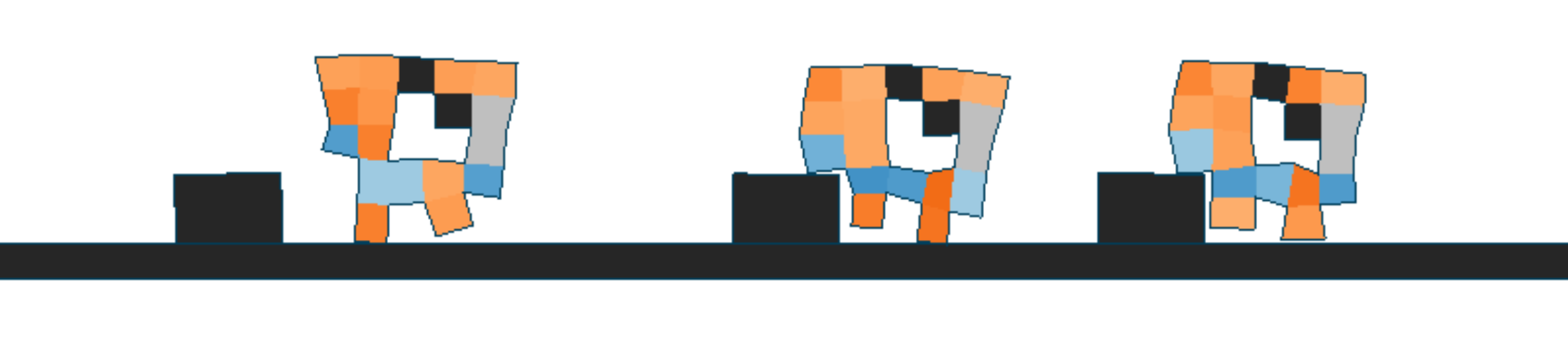}
    \caption{\texttt{Pusher-v1}}
    \label{fig:push_small_rect1}
\end{figure}

In this task the robot pushes/drags a box initialized behind it in the forward direction. This task is \textcolor{cadmiumorange}{\textbf{medium}}.

Let the robot object be $r$ and the box object the robot is trying to push be $b$. The observation space has dimension $\mathcal{S} \in R^{n + 6}$, where $n$ is the number of point masses in object $r$, and is formed by concatenating vectors 
$$\text{\vel{b}{}},\, \text{\pos{b}{}} - \text{\pos{r}{}},\, \text{\vel{r}{}, \, \text{\rel{r}}}$$
with lengths $2$, $n$, $2$, and $2$, respectively. The reward $R = R_1 + R_2$ is the sum of several components. 
$$R_1 = 0.5 \cdot \Delta\text{\pos{r}{x}}  +  0.75 \cdot \Delta\text{\pos{b}{x}}$$
which rewards the robot and box for moving in the positive $x$-direction.
$$R_2 = -\Delta|\text{\pos{b}{x}} - \text{\pos{r}{x}}|$$
which penalizes the robot and box for separating in the $x$-direction.

This environment runs for $600$ steps. The robot also receives a one-time reward of $1$ for reaching the end of the terrain.

\subsubsection{Thrower-v0}

\begin{figure}[H]
    \centering
    \includegraphics[width=0.8\textwidth]{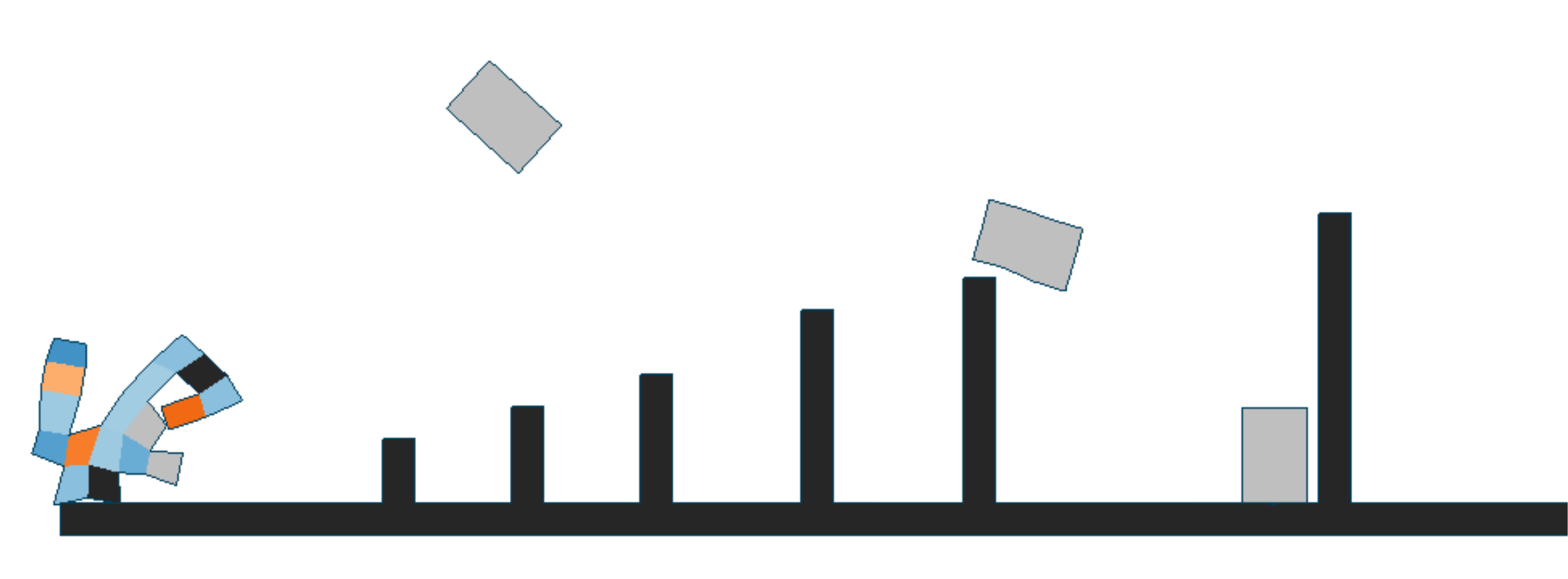}
    \caption{\texttt{Thrower-v0}}
    \label{fig:throw_small_rect}
\end{figure}

In this task the robot throws a box initialized on top of it. This task is \textcolor{cadmiumorange}{\textbf{medium}}.

Let the robot object be $r$ and the box object the robot is trying to throw be $b$. The observation space has dimension $\mathcal{S} \in R^{n + 6}$, where $n$ is the number of point masses in object $r$, and is formed by concatenating vectors 
$$\text{\vel{b}{}},\, \text{\pos{b}{}} - \text{\pos{r}{}},\, \text{\vel{r}{}, \, \text{\rel{r}}}$$
with lengths $2$, $n$, $2$, and $2$, respectively. The reward $R = R_1 + R_2$ is the sum of several components. 
$$R_1 = \Delta\text{\pos{b}{x}}$$
which rewards the box for moving in the positive $x$-direction.
$$R_2 = 
\begin{cases} 
    -\Delta\text{\pos{r}{x}} & \mbox{if } \text{\pos{r}{x}} \ge 0 \\
    \Delta\text{\pos{r}{x}} & \text{otherwise} 
\end{cases}$$
which penalizes the robot for moving too far from $x=0$ when throwing the box.

This environment runs for $300$ steps.

\subsubsection{Catcher-v0}

\begin{figure}[H]
    \centering
    \includegraphics[width=0.8\textwidth]{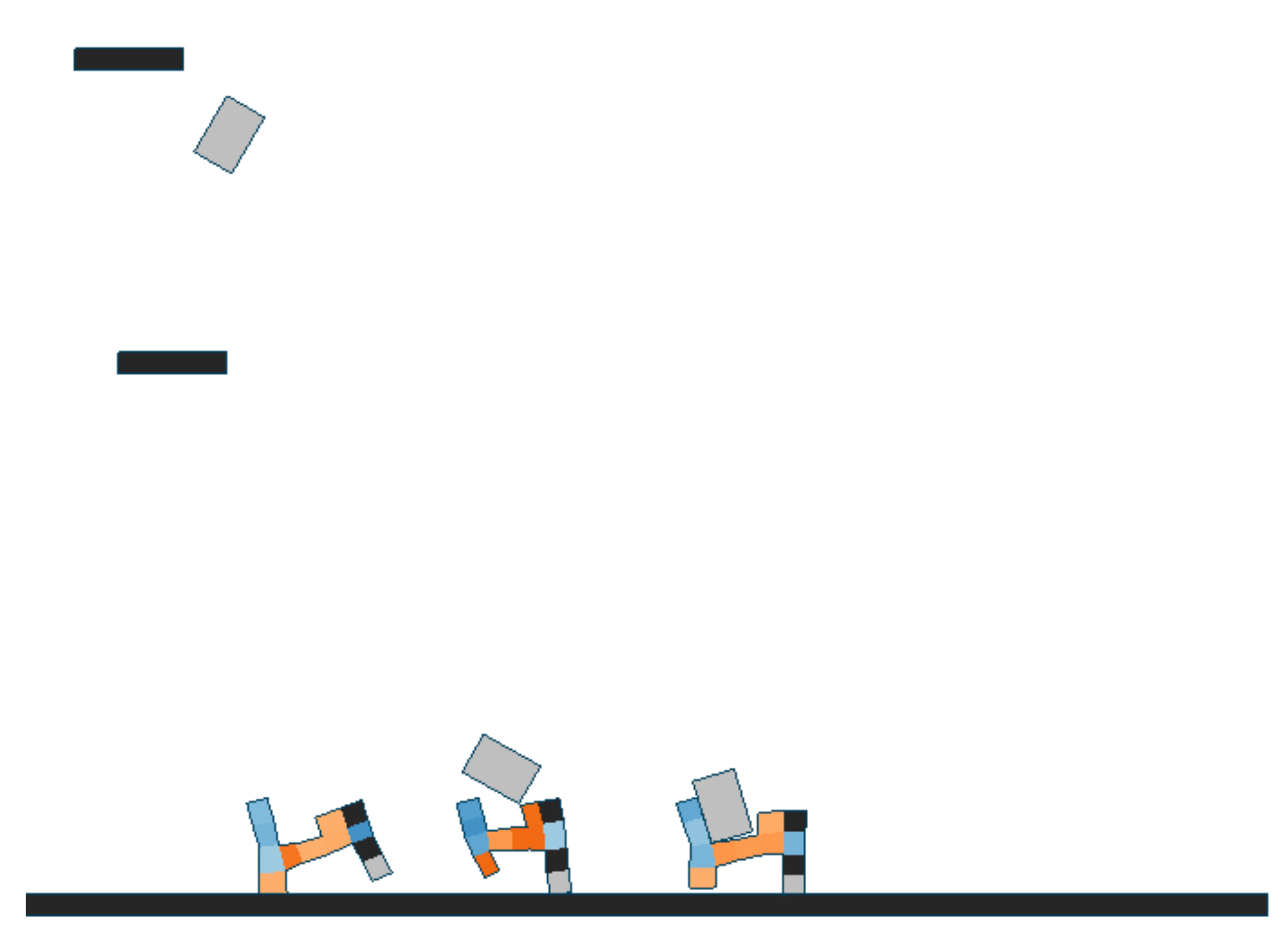}
    \caption{\texttt{Catcher-v0}}
    \label{fig:catch_small_rect}
\end{figure}

In this task the robot catches a fast-moving, rotating box.  This task is \textcolor{cadmiumred}{\textbf{hard}}.

Let the robot object be $r$ and the box object the robot is trying to throw be $b$. The observation space has dimension $\mathcal{S} \in R^{n + 7}$, where $n$ is the number of point masses in object $r$, and is formed by concatenating vectors 
$$\text{\pos{b}{}} - \text{\pos{r}{}},\, \text{\vel{r}{},\, \text{\vel{b}{}},\, \text{\ort{b}},\, \text{\rel{r}}}$$
with lengths $2$, $2$, $2$, $1$, and $n$, respectively. The reward $R = R_1 + R_2$ is the sum of several components. 
$$R_1 = -\Delta|\text{\pos{b}{x}} - \text{\pos{r}{x}}|$$
which rewards the robot for moving to the box in the $x$-direction.
$$R_2 = 
\begin{cases} 
    0 & \mbox{if } \text{\pos{b}{y}} \ge t_y \\
10 \cdot \Delta\text{\pos{b}{y}} & \text{otherwise} 
\end{cases}$$
which penalizes the robot for dropping the box below a threshold height $t_y$. 

This environment runs for $400$ steps.

\subsubsection{BeamToppler-v0}

\begin{figure}[H]
    \centering
    \includegraphics[width=0.8\textwidth]{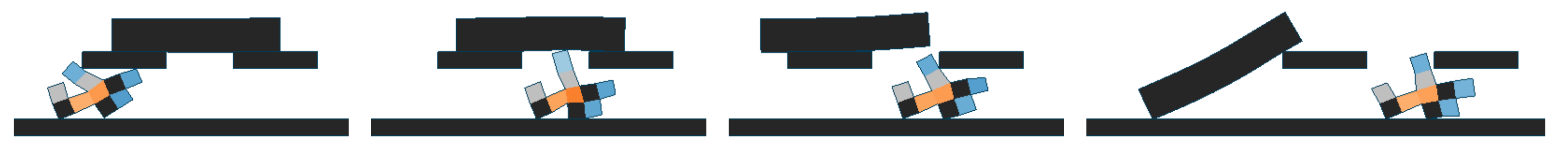}
    \caption{\texttt{BeamToppler-v0}}
    \label{fig:topple_beam}
\end{figure}

In this task the robot knocks over a beam sitting on two pegs from underneath. This task is \textcolor{cadmiumgreen}{\textbf{easy}}.

Let the robot object be $r$ and the beam object the robot is trying to topple be $b$. The observation space has dimension $\mathcal{S} \in R^{n + 7}$, where $n$ is the number of point masses in object $r$, and is formed by concatenating vectors 
$$\text{\pos{b}{}} - \text{\pos{r}{}},\, \text{\vel{r}{},\, \text{\vel{b}{}},\, \text{\ort{b}},\, \text{\rel{r}}}$$
with lengths $2$, $2$, $2$, $1$, and $n$, respectively. The reward $R = R_1 + R_2 + R_3$ is the sum of several components. 
$$R_1 = -\Delta|\text{\pos{b}{x}} - \text{\pos{r}{x}}|$$
which rewards the robot for moving to the beam in the $x$-direction.
$$R_2 = |\Delta\text{\pos{b}{x}}| + 3 \cdot |\Delta\text{\pos{b}{y}}|$$
which rewards the robot for moving the beam. 
$$R_3 = -\Delta\text{\pos{b}{y}}$$
which rewards the robot for making the beam fall. 

This environment runs for $1000$ steps. The robot also receives a one-time reward of $1$ for completing the task.

\subsubsection{BeamSlider-v0}

\begin{figure}[H]
    \centering
    \includegraphics[width=0.8\textwidth]{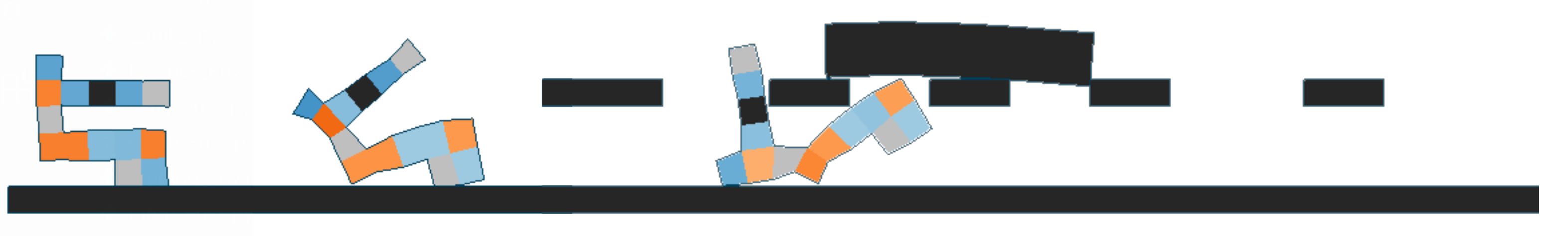}
    \caption{\texttt{BeamSlider-v0}}
    \label{fig:slide_beam}
\end{figure}

In this task the robot slides a beam across a line of pegs from underneath. This task is \textcolor{cadmiumred}{\textbf{hard}}.

Let the robot object be $r$ and the beam object the robot is trying to slide be $b$. The observation space has dimension $\mathcal{S} \in R^{n + 7}$, where $n$ is the number of point masses in object $r$, and is formed by concatenating vectors 
$$\text{\pos{b}{}} - \text{\pos{r}{}},\, \text{\vel{r}{},\, \text{\vel{b}{}},\, \text{\ort{b}},\, \text{\rel{r}}}$$
with lengths $2$, $2$, $2$, $1$, and $n$, respectively. The reward $R = R_1 + R_2$ is the sum of several components. 
$$R_1 = -\Delta|\text{\pos{b}{x}} - \text{\pos{r}{x}}|$$
which rewards the robot for moving to the beam in the $x$-direction.
$$R_2 = \Delta\text{\pos{b}{x}}$$
which rewards the robot for moving the beam in the positive $x$-direction.

This environment runs for $1000$ steps.

\subsubsection{Lifter-v0}

\begin{figure}[H]
    \centering
    \includegraphics[width=0.8\textwidth]{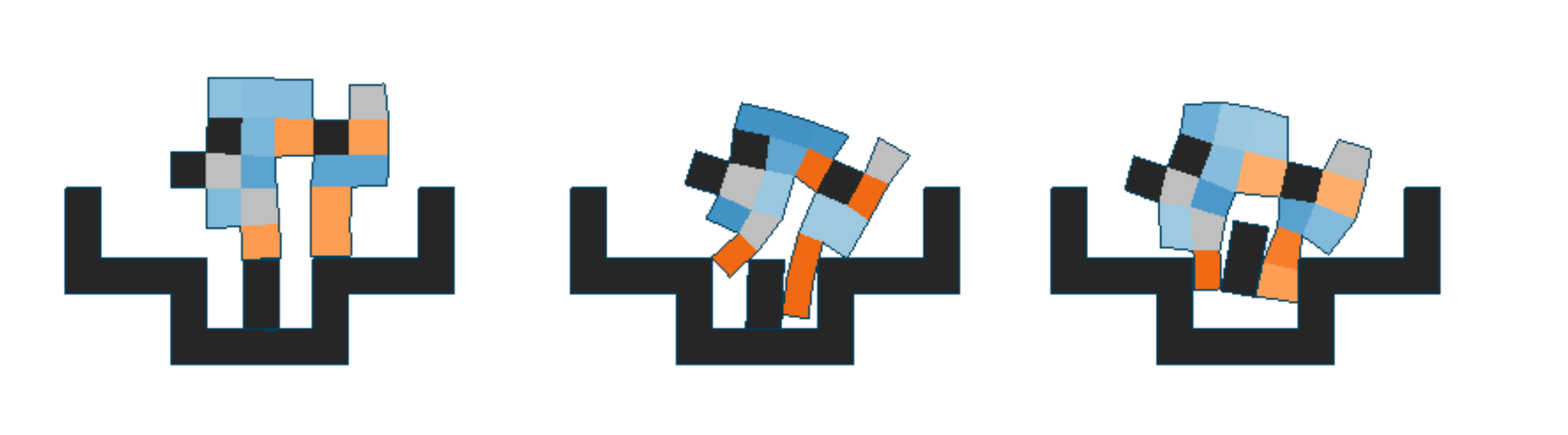}
    \caption{\texttt{Lifter-v0}}
    \label{fig:lift_small_rect}
\end{figure}

In this task the robot lifts a box from out of a hole. This task is \textcolor{cadmiumred}{\textbf{hard}}.

Let the robot object be $r$ and the box object the robot is trying to lift be $b$. The observation space has dimension $\mathcal{S} \in R^{n + 7}$, where $n$ is the number of point masses in object $r$, and is formed by concatenating vectors 
$$\text{\pos{b}{}} - \text{\pos{r}{}},\, \text{\vel{r}{},\, \text{\vel{b}{}},\, \text{\ort{b}},\, \text{\rel{r}}}$$
with lengths $2$, $2$, $2$, $1$, and $n$, respectively. The reward $R = R_1 + R_2 + R_3$ is the sum of several components. 
$$R_1 = 10 \cdot \Delta\text{\pos{b}{y}}$$
which rewards the robot for moving the beam in the positive $y$-direction.
$$R_2 = -10 \cdot \Delta|g_x - \text{\pos{b}{x}}|$$
which penalizes the robot for moving the box away from a goal $x$-position, $g_x$. This ensures that the robot lifts the box straight up.
$$R_3 = 
\begin{cases} 
    0 & \mbox{if } \text{\pos{r}{y}} \ge t_y \\
20 \cdot \Delta\text{\pos{r}{y}} & \text{otherwise} 
\end{cases}$$
which penalizes the robot for falling below a threshold height $t_y$ (at which point the robot has fallen into the hole). 

This environment runs for $300$ steps.

\subsection{Climbing tasks}

\subsubsection{Climber-v0}

\begin{figure}[H]
    \centering
    \includegraphics[width=0.1\textwidth]{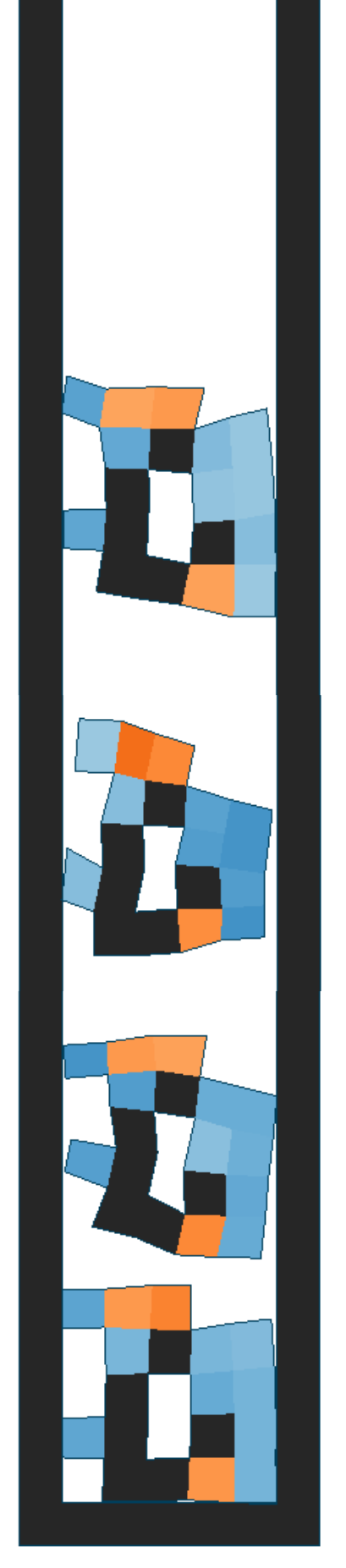}
    \caption{\texttt{Climber-v0}}
    \label{fig:climb}
\end{figure}

In this task the robot climbs as high as possible through a flat, vertical channel. This task is \textcolor{cadmiumorange}{\textbf{medium}}.

Let the robot object be $r$. The observation space has dimension $\mathcal{S} \in R^{n + 2}$, where $n$ is the number of point masses in object $r$, and is formed by concatenating vectors 
$$\text{\vel{r}{}, \rel{r}}$$
with lengths $2$ and $n$, respectively. The reward $R$ is 
$$R = \Delta\text{\pos{r}{y}}$$
which rewards the robot for moving in the positive $y$-direction.

This environment runs for $400$ steps. The robot also receives a one-time reward of $1$ for reaching the end of the terrain.

\subsubsection{Climber-v1}

\begin{figure}[H]
    \centering
    \includegraphics[width=0.1\textwidth]{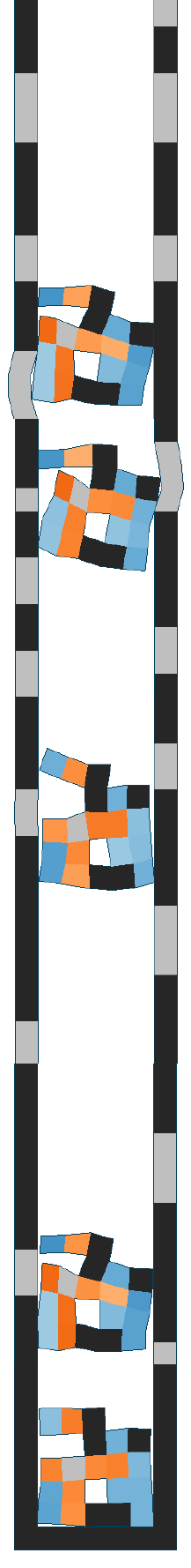}
    \caption{\texttt{Climber-v1}}
    \label{fig:climb1}
\end{figure}

In this task the robot climbs as high as possible through a vertical channel made of mixed rigid and soft materials. This task is \textcolor{cadmiumorange}{\textbf{medium}}.

Let the robot object be $r$. The observation space has dimension $\mathcal{S} \in R^{n + 2}$, where $n$ is the number of point masses in object $r$, and is formed by concatenating vectors 
$$\text{\vel{r}{}, \rel{r}}$$
with lengths $2$ and $n$, respectively. The reward $R$ is 
$$R = \Delta\text{\pos{r}{y}}$$
which rewards the robot for moving in the positive $y$-direction.

This environment runs for $600$ steps. The robot also receives a one-time reward of $1$ for reaching the end of the terrain. 

\subsubsection{Climber-v2}

\begin{figure}[H]
    \centering
    \includegraphics[width=0.1\textwidth]{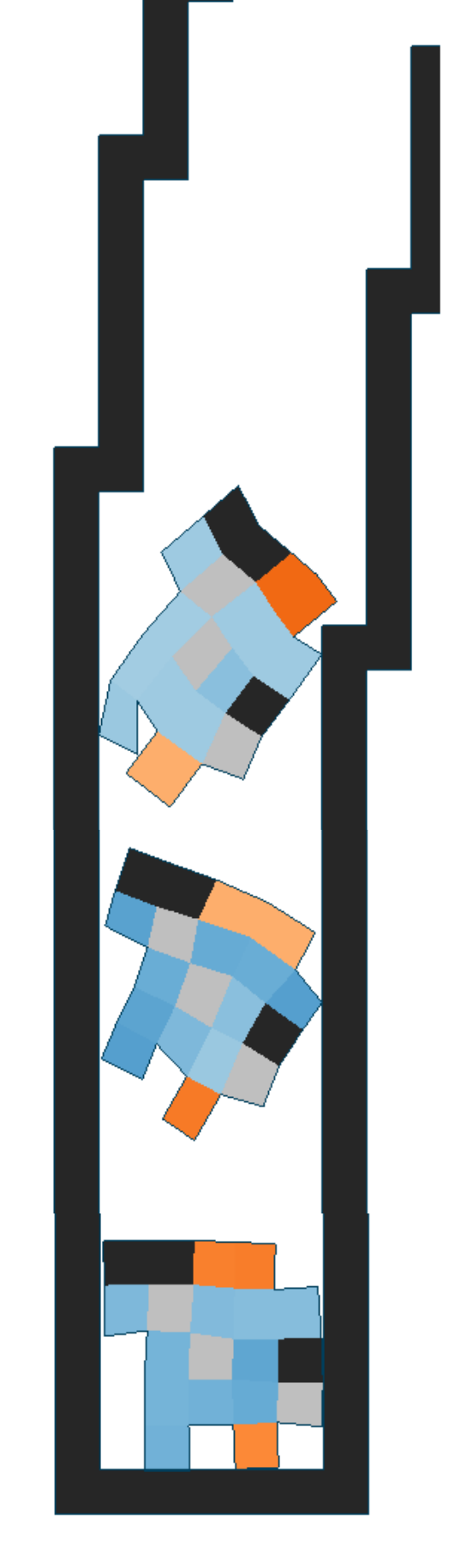}
    \caption{\texttt{Climber-v2}}
    \label{fig:climb3}
\end{figure}

In this task the robot climbs as high as possible through a narrow stepwise channel. This task is \textcolor{cadmiumred}{\textbf{hard}}.

Let the robot object be $r$. The observation space has dimension $\mathcal{S} \in R^{n + 10}$, where $n$ is the number of point masses in object $r$, and is formed by concatenating vectors 
$$\text{\vel{r}{}, \ort{r}, \rel{r}, \abo{r}{3}}$$
with lengths $2$, $1$, $n$, and $7$ respectively. The reward $R$ is 
$$R = \Delta\text{\pos{r}{y}} + 0.2 \cdot \Delta\text{\pos{r}{x}}$$
which rewards the robot for moving in the positive $y$-direction and positive $x$-direction.

This environment runs for $1000$ steps.

\subsection{Forward locomotion tasks}

\subsubsection{UpStepper-v0}

\begin{figure}[H]
    \centering
    \includegraphics[width=0.8\textwidth]{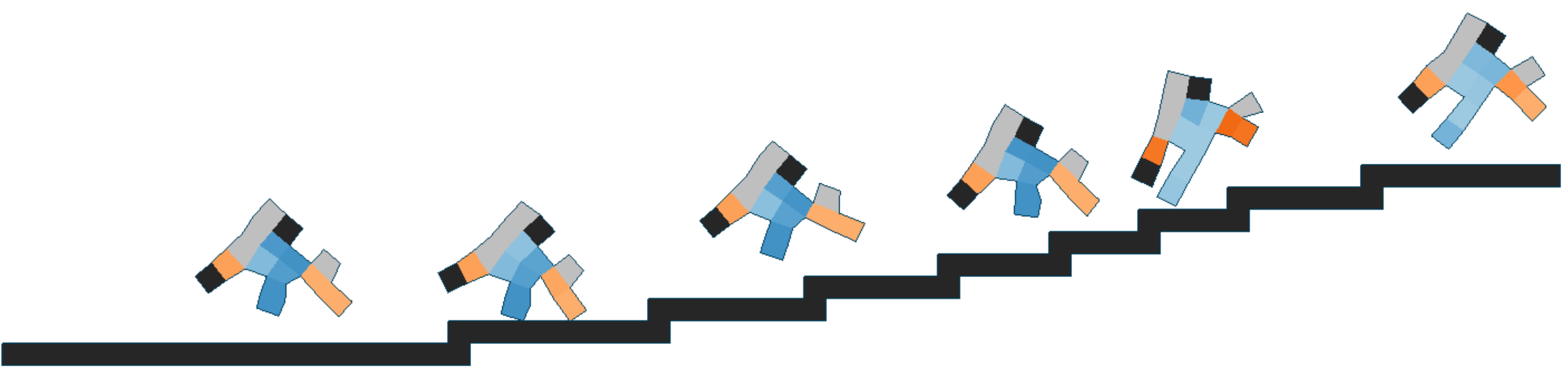}
    \caption{\texttt{UpStepper-v0}}
    \label{fig:steps_up}
\end{figure}

In this task the robot climbs up stairs of varying lengths. This task is \textcolor{cadmiumorange}{\textbf{medium}}.

Let the robot object be $r$. The observation space has dimension $\mathcal{S} \in R^{n + 14}$, where $n$ is the number of point masses in object $r$, and is formed by concatenating vectors 
$$\text{\vel{r}{}, \ort{r}, \rel{r}, \bel{r}{5}}$$
with lengths $2$, $1$, $n$, and $11$ respectively. The reward $R$ is 
$$R = \Delta\text{\pos{r}{x}}$$
which rewards the robot for moving in the positive $x$-direction. The robot also receives a one-time reward of $2$ for reaching the end of the terrain, and a one-time penalty of $-3$ for rotating more than $75$ degrees from its originally orientation in either direction (after which the environment resets).

This environment runs for $600$ steps.

\subsubsection{DownStepper-v0}

\begin{figure}[H]
    \centering
    \includegraphics[width=0.8\textwidth]{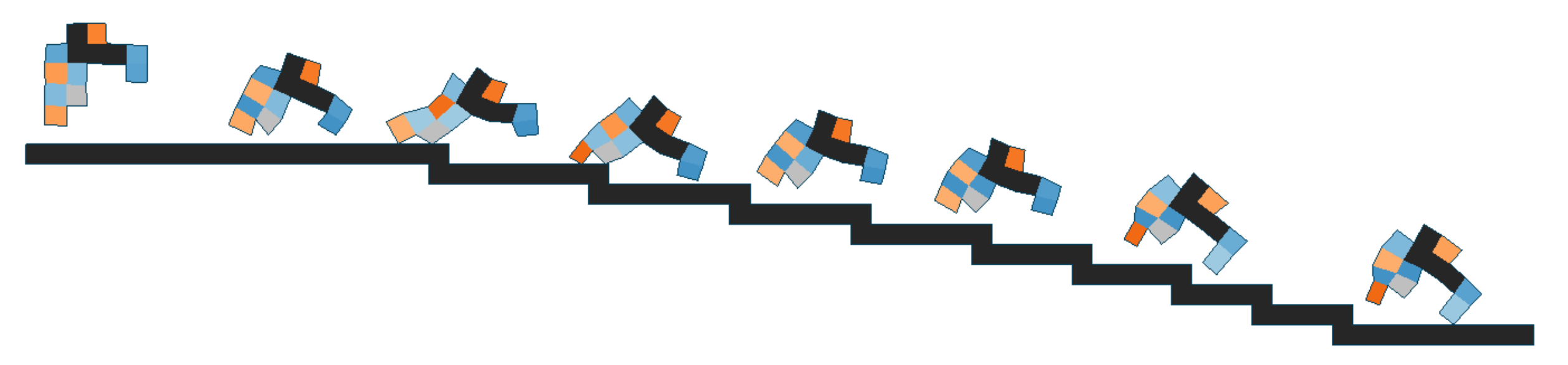}
    \caption{\texttt{DownStepper-v0}}
    \label{fig:steps_down}
\end{figure}

In this task the robot climbs down stairs of varying lengths. This task is \textcolor{cadmiumgreen}{\textbf{easy}}.

Let the robot object be $r$. The observation space has dimension $\mathcal{S} \in R^{n + 14}$, where $n$ is the number of point masses in object $r$, and is formed by concatenating vectors 
$$\text{\vel{r}{}, \ort{r}, \rel{r}, \bel{r}{5}}$$
with lengths $2$, $1$, $n$, and $11$ respectively. The reward $R$ is 
$$R = \Delta\text{\pos{r}{x}}$$
which rewards the robot for moving in the positive $x$-direction. The robot also receives a one-time reward of $2$ for reaching the end of the terrain, and a one-time penalty of $-3$ for rotating more than $90$ degrees from its originally orientation in either direction (after which the environment resets).

This environment runs for $500$ steps.

\subsubsection{ObstacleTraverser-v0}

\begin{figure}[H]
    \centering
    \includegraphics[width=0.8\textwidth]{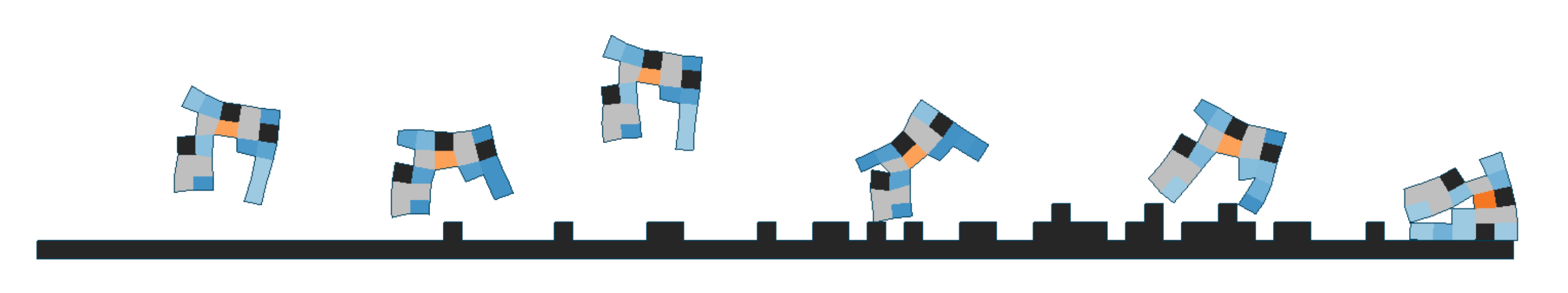}
    \caption{\texttt{ObstacleTraverser-v0}}
    \label{fig:walking_bumpy}
\end{figure}

In this task the robot walks across terrain that gets increasingly more bumpy. This task is \textcolor{cadmiumorange}{\textbf{medium}}.

Let the robot object be $r$. The observation space has dimension $\mathcal{S} \in R^{n + 14}$, where $n$ is the number of point masses in object $r$, and is formed by concatenating vectors 
$$\text{\vel{r}{}, \ort{r}, \rel{r}, \bel{r}{5}}$$
with lengths $2$, $1$, $n$, and $11$ respectively. The reward $R$ is 
$$R = \Delta\text{\pos{r}{x}}$$
which rewards the robot for moving in the positive $x$-direction. The robot also receives a one-time reward of $2$ for reaching the end of the terrain, and a one-time penalty of $-3$ for rotating more than $90$ degrees from its originally orientation in either direction (after which the environment resets).

This environment runs for $1000$ steps.

\subsubsection{ObstacleTraverser-v1}

\begin{figure}[H]
    \centering
    \includegraphics[width=0.8\textwidth]{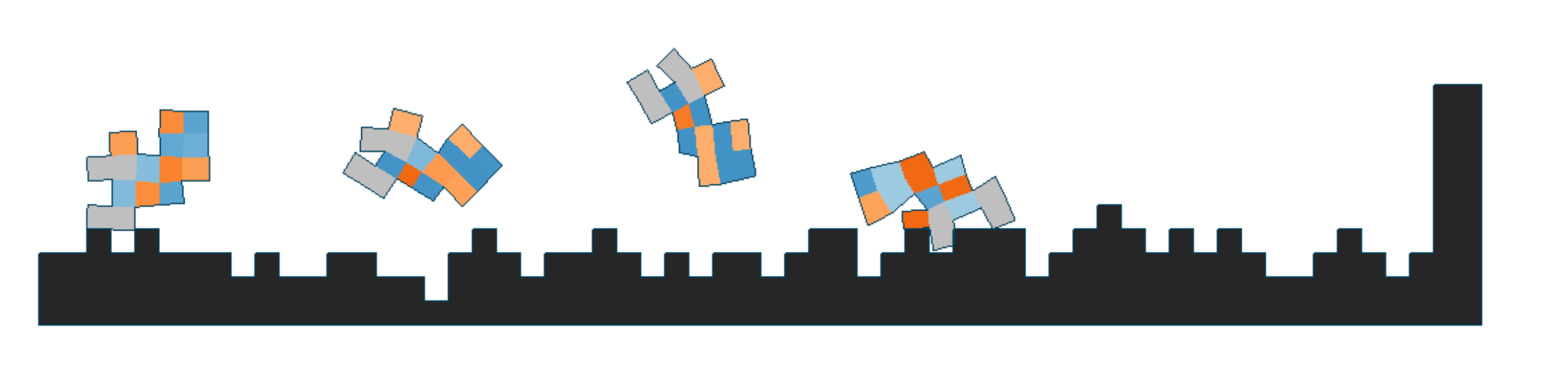}
    \caption{\texttt{ObstacleTraverser-v1}}
    \label{fig:walking_bumpy1}
\end{figure}

In this task the robot walks through very bumpy terrain. This task is \textcolor{cadmiumred}{\textbf{hard}}.

Let the robot object be $r$. The observation space has dimension $\mathcal{S} \in R^{n + 14}$, where $n$ is the number of point masses in object $r$, and is formed by concatenating vectors 
$$\text{\vel{r}{}, \ort{r}, \rel{r}, \bel{r}{5}}$$
with lengths $2$, $1$, $n$, and $11$ respectively. The reward $R$ is 
$$R = \Delta\text{\pos{r}{x}}$$
which rewards the robot for moving in the positive $x$-direction. The robot also receives a one-time reward of $2$ for reaching the end of the terrain (after which the environment resets).

This environment runs for $1000$ steps.

\subsubsection{Hurdler-v0}

\begin{figure}[H]
    \centering
    \includegraphics[width=0.8\textwidth]{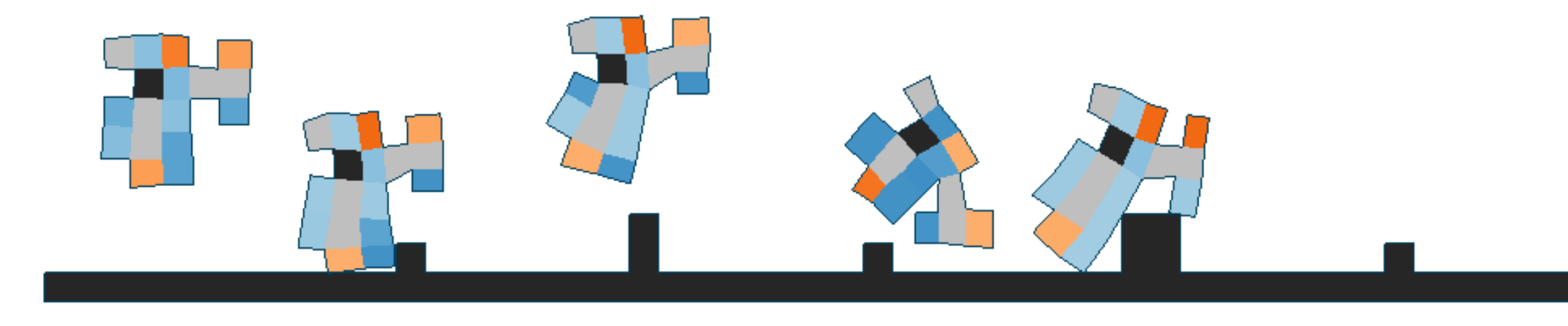}
    \caption{\texttt{Hurdler-v0}}
    \label{fig:vertical_barrier}
\end{figure}

In this task the robot walks across terrain with tall obstacles. This task is \textcolor{cadmiumred}{\textbf{hard}}.

Let the robot object be $r$. The observation space has dimension $\mathcal{S} \in R^{n + 14}$, where $n$ is the number of point masses in object $r$, and is formed by concatenating vectors 
$$\text{\vel{r}{}, \ort{r}, \rel{r}, \bel{r}{5}}$$
with lengths $2$, $1$, $n$, and $11$ respectively. The reward $R$ is 
$$R = \Delta\text{\pos{r}{x}}$$
which rewards the robot for moving in the positive $x$-direction. The robot also receives a one-time penalty of $-3$ for rotating more than $90$ degrees from its originally orientation in either direction (after which the environment resets).

This environment runs for $1000$ steps.

\subsubsection{PlatformJumper-v0}

\begin{figure}[H]
    \centering
    \includegraphics[width=0.8\textwidth]{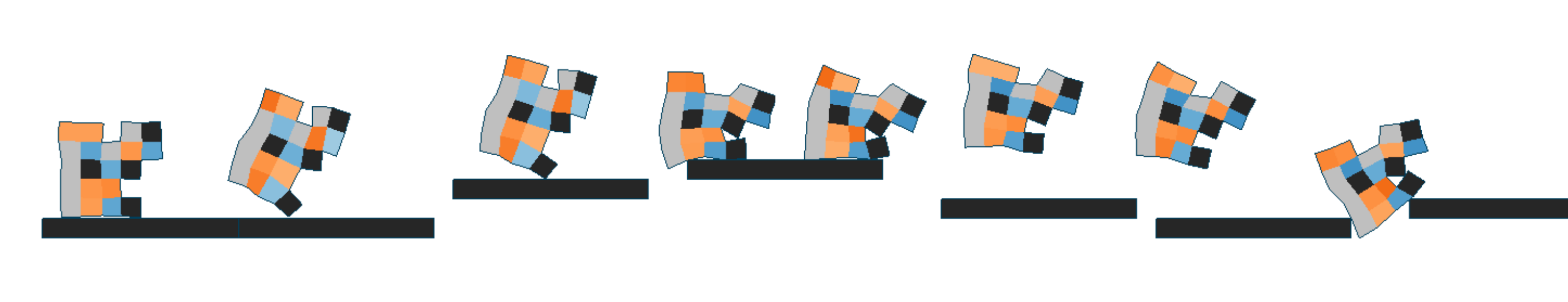}
    \caption{\texttt{PlatformJumper-v0}}
    \label{fig:floating_platform}
\end{figure}

In this task the robot traverses a series of floating platforms at different heights. This task is \textcolor{cadmiumred}{\textbf{hard}}.

Let the robot object be $r$. The observation space has dimension $\mathcal{S} \in R^{n + 14}$, where $n$ is the number of point masses in object $r$, and is formed by concatenating vectors 
$$\text{\vel{r}{}, \ort{r}, \rel{r}, \bel{r}{5}}$$
with lengths $2$, $1$, $n$, and $11$ respectively. The reward $R$ is 
$$R = \Delta\text{\pos{r}{x}}$$
which rewards the robot for moving in the positive $x$-direction. The robot also receives a one-time penalty of $-3$ for rotating more than $90$ degrees from its originally orientation in either direction or for falling off the platforms (after which the environment resets).

This environment runs for $1000$ steps.

\subsubsection{GapJumper-v0}

\begin{figure}[H]
    \centering
    \includegraphics[width=0.8\textwidth]{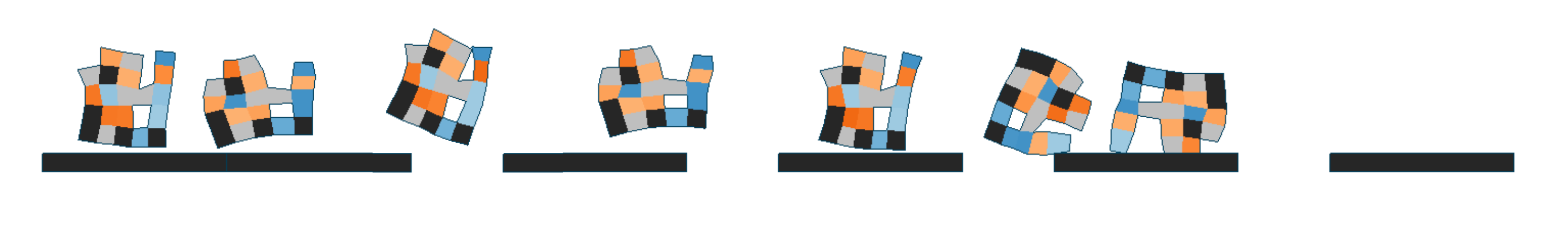}
    \caption{\texttt{GapJumper-v0}}
    \label{fig:gaps}
\end{figure}

In this task the robot traverses a series of spaced-out floating platforms all at the same height. This task is \textcolor{cadmiumred}{\textbf{hard}}.

Let the robot object be $r$. The observation space has dimension $\mathcal{S} \in R^{n + 14}$, where $n$ is the number of point masses in object $r$, and is formed by concatenating vectors 
$$\text{\vel{r}{}, \ort{r}, \rel{r}, \bel{r}{5}}$$
with lengths $2$, $1$, $n$, and $11$ respectively. The reward $R$ is 
$$R = \Delta\text{\pos{r}{x}}$$
which rewards the robot for moving in the positive $x$-direction. The robot also receives a one-time penalty of $-3$ for falling off the platforms (after which the environment resets).

This environment runs for $1000$ steps.

\subsubsection{Traverser-v0}

\begin{figure}[H]
    \centering
    \includegraphics[width=0.8\textwidth]{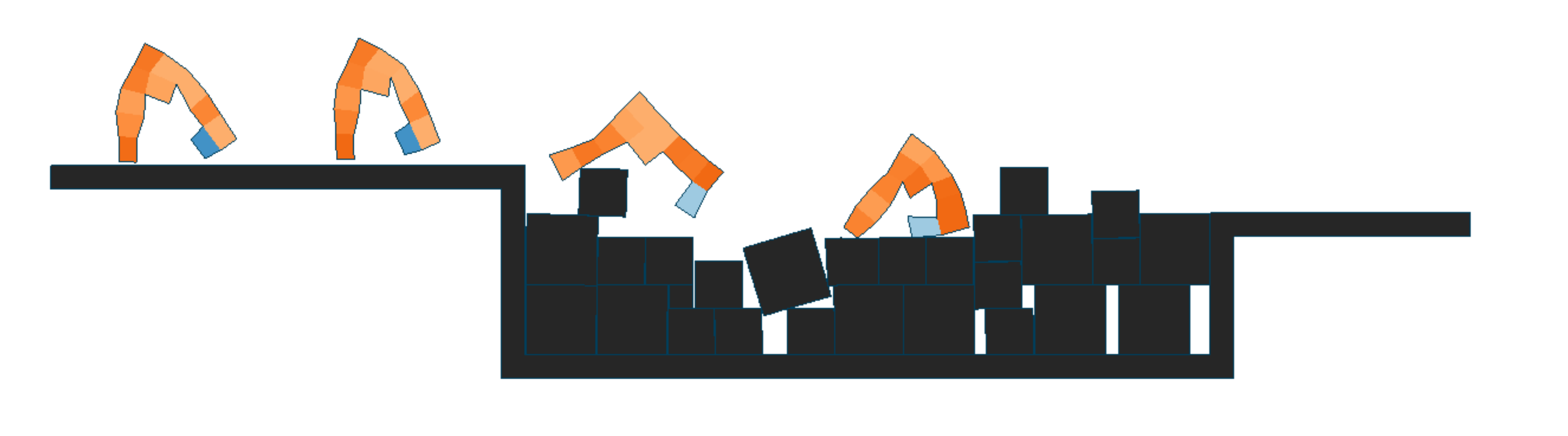}
    \caption{\texttt{Traverser-v0}}
    \label{fig:block_soup}
\end{figure}

In this task the robot traverses a pit of rigid blocks to get to the other side without sinking into the pit. This task is \textcolor{cadmiumred}{\textbf{hard}}.

Let the robot object be $r$. The observation space has dimension $\mathcal{S} \in R^{n + 14}$, where $n$ is the number of point masses in object $r$, and is formed by concatenating vectors 
$$\text{\vel{r}{}, \ort{r}, \rel{r}, \bel{r}{5}}$$
with lengths $2$, $1$, $n$, and $11$ respectively. The reward $R$ is 
$$R = \Delta\text{\pos{r}{x}}$$
which rewards the robot for moving in the positive $x$-direction. The robot also receives a one-time reward off $2$ for reaching the end of the terrain (after which the environment resets).

This environment runs for $1000$ steps.

\subsubsection{CaveCrawler-v0}

\begin{figure}[H]
    \centering
    \includegraphics[width=0.8\textwidth]{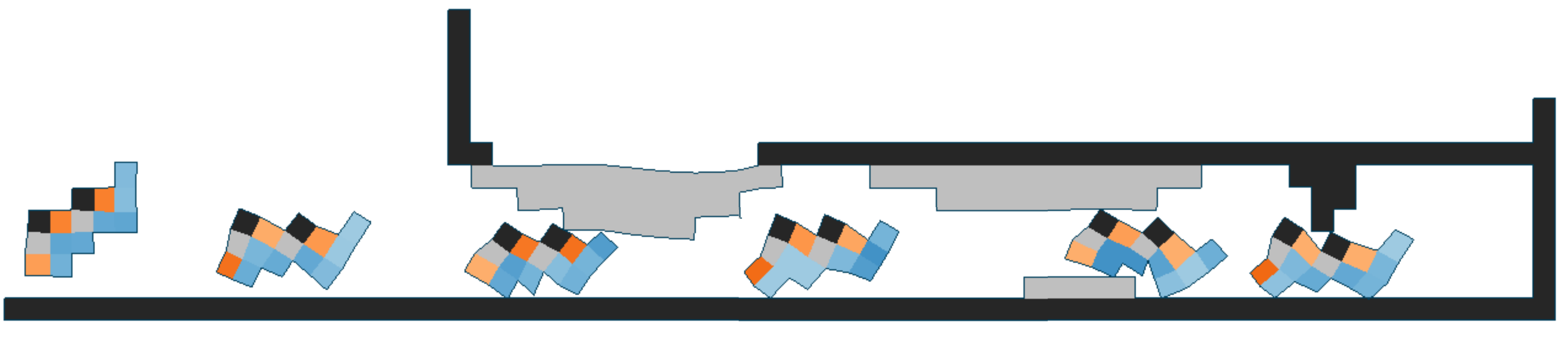}
    \caption{\texttt{CaveCrawler-v0}}
    \label{fig:duck}
\end{figure}

In this task the robot squeezes through caves and low-hanging obstacles. This task is \textcolor{cadmiumorange}{\textbf{medium}}.

Let the robot object be $r$. The observation space has dimension $\mathcal{S} \in R^{n + 24}$, where $n$ is the number of point masses in object $r$, and is formed by concatenating vectors 
$$\text{\vel{r}{}, \rel{r}, \bel{r}{5}, \abo{r}{5}}$$
with lengths $2$, $n$, $11$, and $11$ respectively. The reward $R$ is 
$$R = \Delta\text{\pos{r}{x}}$$
which rewards the robot for moving in the positive $x$-direction. The robot also receives a one-time reward off $1$ for reaching the end of the terrain (after which the environment resets).

This environment runs for $1000$ steps.

\subsection{Shape change tasks}

\subsubsection{AreaMaximizer-v0}

\begin{figure}[H]
    \centering
    \includegraphics[width=0.8\textwidth]{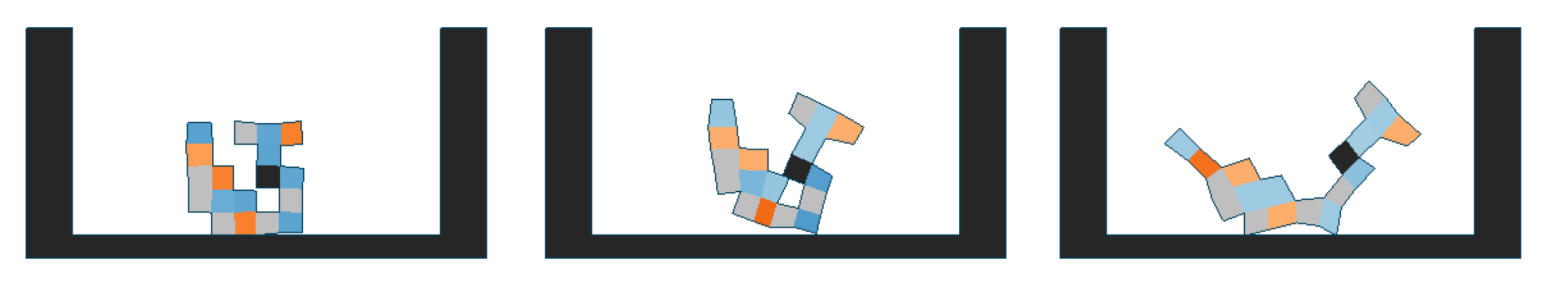}
    \caption{\texttt{AreaMaximizer-v0}}
    \label{fig:max_shape}
\end{figure}

In this task the robot grows to occupy the largest possible surface area. This task is \textcolor{cadmiumgreen}{\textbf{easy}}.

Let the robot object be $r$. The observation space has dimension $\mathcal{S} \in R^{n}$, where $n$ is the number of point masses in object $r$, and is simply the vector 
$$\text{\rel{r}}$$
with length $n$. Let $a^r$ be the area of the convex hull formed by the point masses of $r$. The reward $R$ is 
$$R = \Delta{a^r}$$
which rewards the robot for growing.

This environment runs for $600$ steps.

\subsubsection{AreaMinimizer-v0}

\begin{figure}[H]
    \centering
    \includegraphics[width=0.6\textwidth]{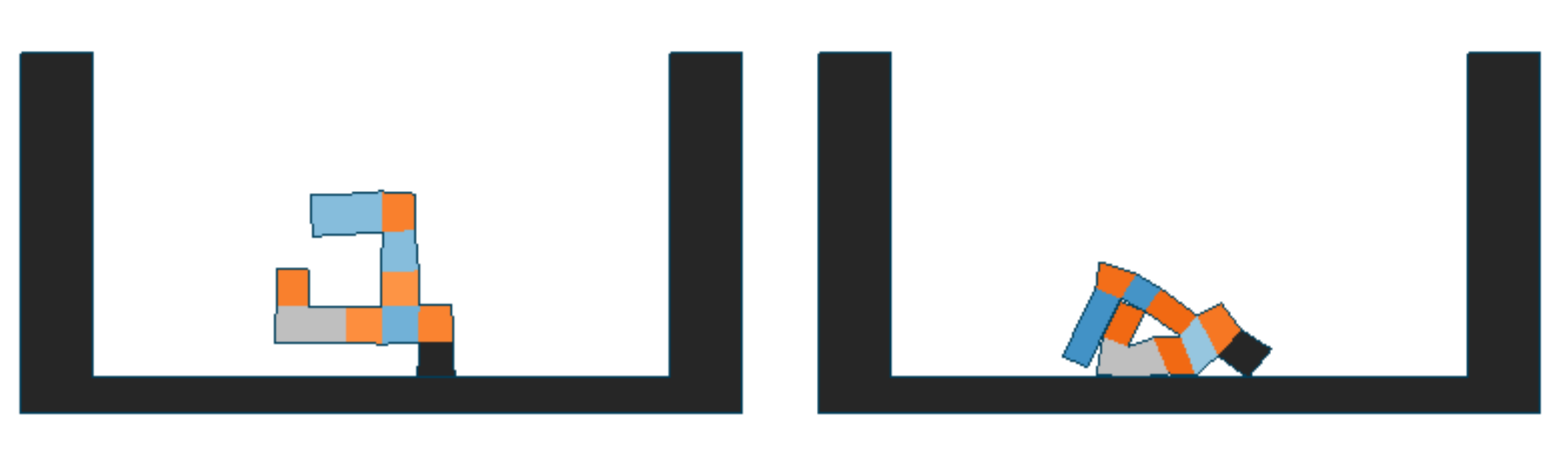}
    \caption{\texttt{AreaMinimizer-v0}}
    \label{fig:min_shape}
\end{figure}

In this task the robot shrinks to occupy the smallest possible surface area. This task is \textcolor{cadmiumorange}{\textbf{medium}}.

Let the robot object be $r$. The observation space has dimension $\mathcal{S} \in R^{n}$, where $n$ is the number of point masses in object $r$, and is simply the vector 
$$\text{\rel{r}}$$
with length $n$. Let $a^r$ be the area of the convex hull formed by the point masses of $r$. The reward $R$ is 
$$R = -\Delta{a^r}$$
which rewards the robot for shrinking.

This environment runs for $600$ steps.

\subsubsection{WingspanMaximizer-v0}

\begin{figure}[H]
    \centering
    \includegraphics[width=0.8\textwidth]{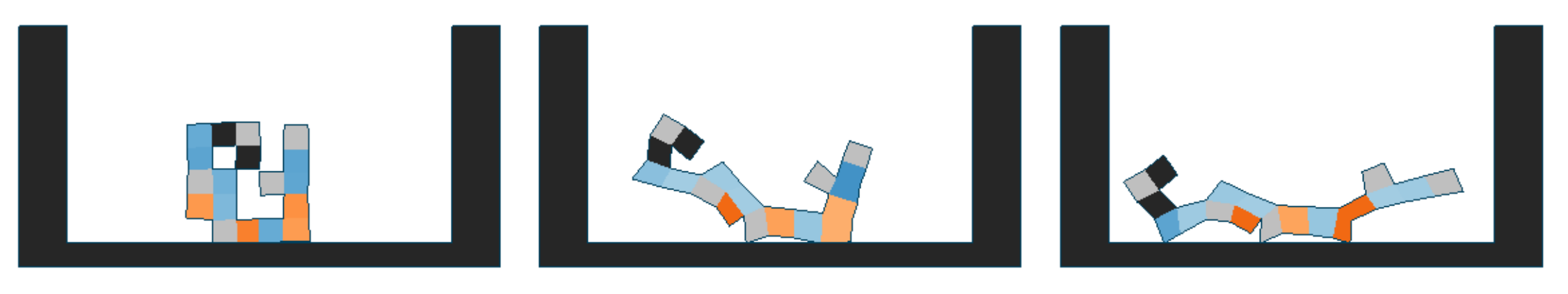}
    \caption{\texttt{WingspanMaximizer-v0}}
    \label{fig:max_x_shape}
\end{figure}

In this task the robot grows to be as wide as possible. This task is \textcolor{cadmiumgreen}{\textbf{easy}}.

Let the robot object be $r$. The observation space has dimension $\mathcal{S} \in R^{n}$, where $n$ is the number of point masses in object $r$, and is simply the vector 
$$\text{\rel{r}}$$
with length $n$. Let $p^i$ be the vector representing the position of point mass $i$ in $r$. The reward $R$ is 
$$R = \Delta{\left[\max_{i}{p^i_x} - \min_{i}{p^i_x} \right]}$$
which rewards the robot for growing in the $x$-direction.

This environment runs for $600$ steps.

\subsubsection{HeightMaximizer-v0}

\begin{figure}[H]
    \centering
    \includegraphics[width=0.8\textwidth]{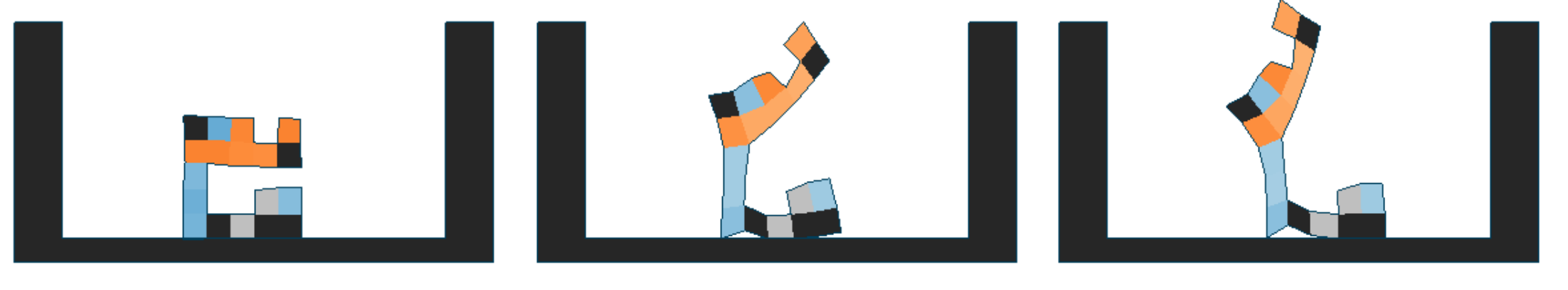}
    \caption{\texttt{HeightMaximizer-v0}}
    \label{fig:max_y_shape}
\end{figure}

In this task the robot grows to be as tall as possible. This task is \textcolor{cadmiumorange}{\textbf{medium}}.

Let the robot object be $r$. The observation space has dimension $\mathcal{S} \in R^{n}$, where $n$ is the number of point masses in object $r$, and is simply the vector 
$$\text{\rel{r}}$$
with length $n$. Let $p^i$ be the vector representing the position of point mass $i$ in $r$. The reward $R$ is 
$$R = \Delta{\left[\max_{i}{p^i_y} - \min_{i}{p^i_y} \right]}$$
which rewards the robot for growing in the $y$-direction.

This environment runs for $500$ steps.

\subsection{Miscellaneous tasks}

\subsubsection{Flipper-v0}

\begin{figure}[H]
    \centering
    \includegraphics[width=0.8\textwidth]{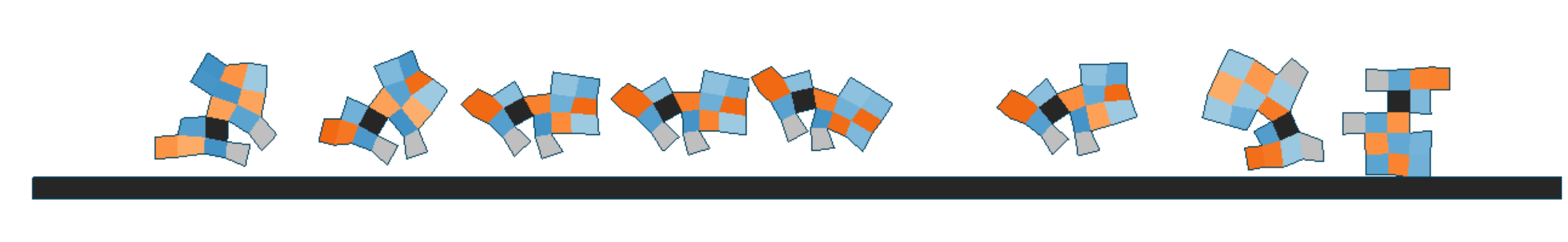}
    \caption{\texttt{Flipper-v0}}
    \label{fig:flipping}
\end{figure}

In this task the robot flips counter-clockwise as many times as possible on flat terrain. This task is \textcolor{cadmiumgreen}{\textbf{easy}}.

Let the robot object be $r$. The observation space has dimension $\mathcal{S} \in R^{n + 1}$, where $n$ is the number of point masses in object $r$, and is formed by concatenating vectors 
$$\text{\ort{r}, \rel{r}}$$
with lengths $1$ and $n$ respectively. The reward $R$ is 
$$R = \Delta\text{\ort{r}}$$
which rewards the robot for rotating counter-clockwise.

This environment runs for $600$ steps.

\subsubsection{Jumper-v0}

\begin{figure}[H]
    \centering
    \includegraphics[width=0.8\textwidth]{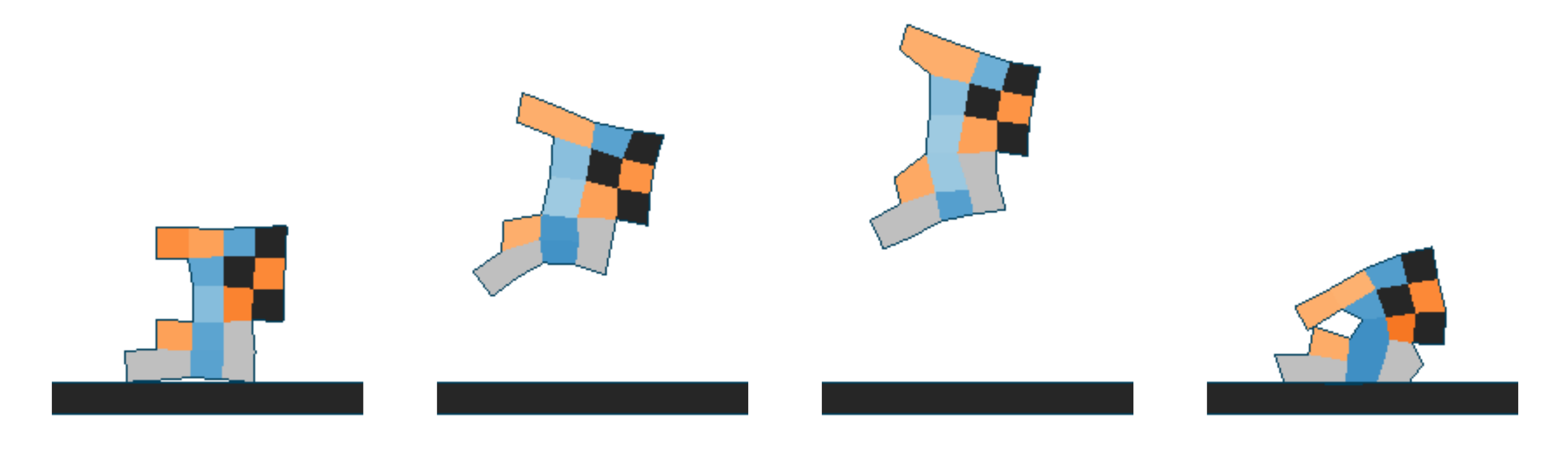}
    \caption{\texttt{Jumper-v0}}
    \label{fig:stationary_jump}
\end{figure}

In this task the robot jumps as high as possible in place on flat terrain. This task is \textcolor{cadmiumgreen}{\textbf{easy}}.

Let the robot object be $r$. The observation space has dimension $\mathcal{S} \in R^{n + 7}$, where $n$ is the number of point masses in object $r$, and is formed by concatenating vectors 
$$\text{\vel{r}{}, \rel{r}, \bel{r}{2}}$$
with lengths $2$, $n$, and $5$ respectively. The reward $R$ is 
$$R = 10\cdot \Delta\text{\pos{r}{y}} - 5 \cdot |\Delta\text{\pos{r}{x}}|$$
which rewards the robot for moving in the positive $y$-direction and penalizes the robot for any motion in the $x$-direction.

This environment runs for $500$ steps.

\subsubsection{Balancer-v0}

\begin{figure}[H]
    \centering
    \includegraphics[width=0.8\textwidth]{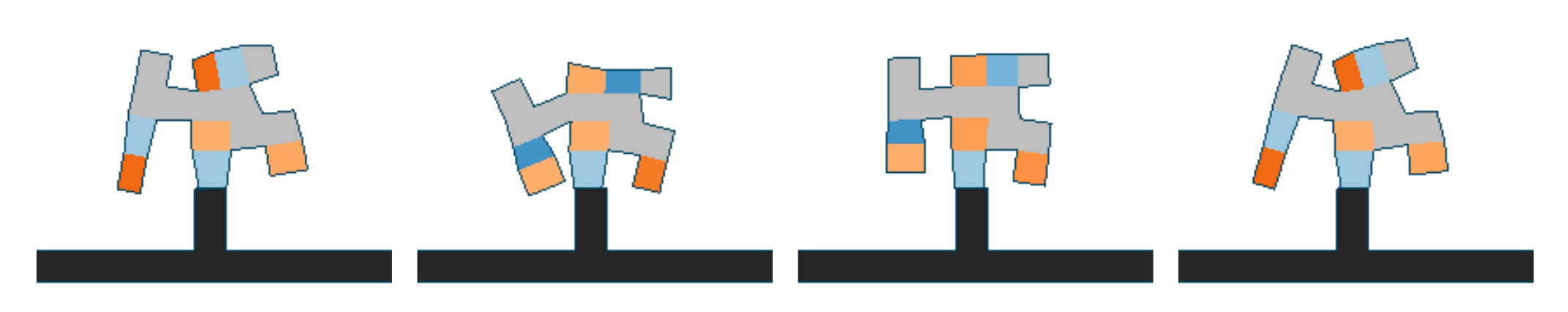}
    \caption{\texttt{Balancer-v0}}
    \label{fig:balance}
\end{figure}

In this task the robot is initialized on top of a thin pole and balances on it. This task is \textcolor{cadmiumgreen}{\textbf{easy}}.

Let the robot object be $r$. We achieve the described behavior by setting a goal position - $g_x$ and $g_y$ - for the robot located on top of the pole. The observation space has dimension $\mathcal{S} \in R^{n + 2}$, where $n$ is the number of point masses in object $r$, and is formed by concatenating vectors 
$$g_x - \text{\pos{r}{x}},\, g_y - \text{\pos{r}{y}},\, \text{\rel{r}}$$
with lengths $1$, $1$, and $n$ respectively. The reward $R$ is 
$$R = -\Delta|g_x - \text{\pos{b}{x}}|-\Delta|g_y - \text{\pos{b}{y}}|$$
which rewards the robot for moving towards the goal in the $x$ and $y$ directions

This environment runs for $600$ steps.

\subsubsection{Balancer-v1}

\begin{figure}[H]
    \centering
    \includegraphics[width=0.8\textwidth]{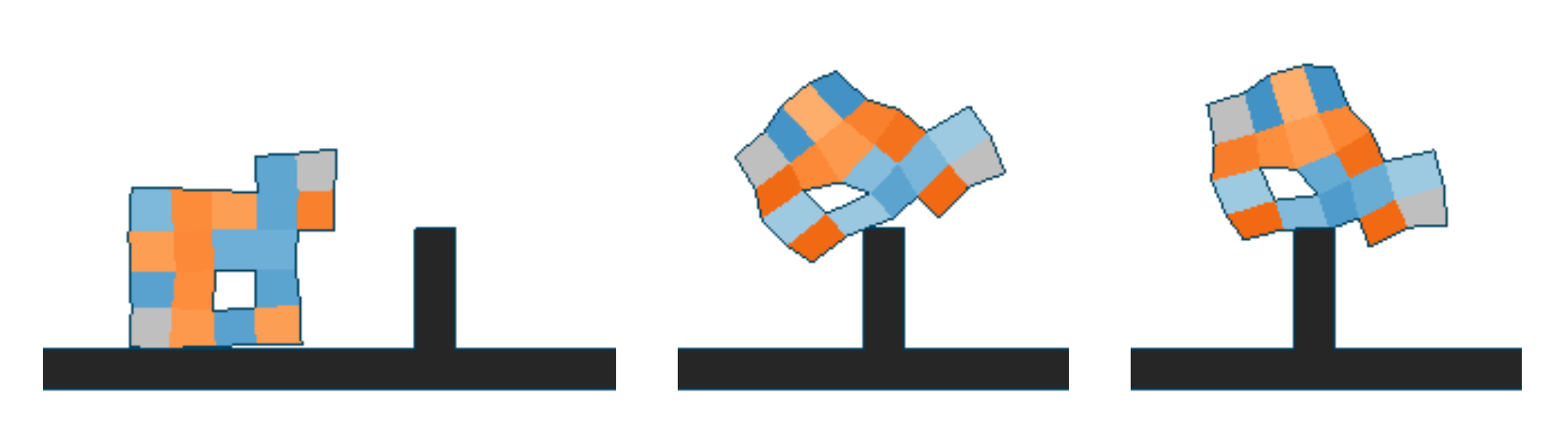}
    \caption{\texttt{Balancer-v1}}
    \label{fig:balance1}
\end{figure}

In this task the robot is initialized next to a thin pole. The robot jumps on the pole and balances on it. This task is \textcolor{cadmiumorange}{\textbf{medium}}.

Let the robot object be $r$. We achieve the described behavior by setting a goal position - $g_x$ and $g_y$ - for the robot located on top of the pole. The observation space has dimension $\mathcal{S} \in R^{n + 2}$, where $n$ is the number of point masses in object $r$, and is formed by concatenating vectors 
$$g_x - \text{\pos{r}{x}},\, g_y - \text{\pos{r}{y}},\, \text{\rel{r}}$$
with lengths $1$, $1$, and $n$ respectively. The reward $R$ is 
$$R = -\Delta|g_x - \text{\pos{b}{x}}|-\Delta|g_y - \text{\pos{b}{y}}|$$
which rewards the robot for moving towards the goal in the $x$ and $y$ directions

This environment runs for $600$ steps.

\section{Optimization methods}
\label{supp:optim}

In this section, we extend the description of optimization methods that we describe in Section \ref{sec:algorithm} of the main paper and also provide detailed pseudocode.


\subsection{Genetic algorithm (GA)}

We implement a simple GA using elitism selection and a simple mutation strategy to evolve the population of robot designs. The selection keeps the top $x$\% of the robots from the current population as survivors and discards the rest, and the mutation can randomly change each voxel of the robot with certain probability. Crossover is not implemented for simplicity, but GA can potentially perform better with carefully designed crossover operators. See Algorithm \ref{alg:ga} for more details in each generation of GA.

\begin{algorithm}
\caption{Genetic algorithm (per generation)}
    \label{alg:ga}
    \begin{algorithmic}
        \STATE \textbf{Inputs:} History data $S$, population size $p$, current generation number $N_{cur}$, max generation number $N_{max}$.
        \STATE \textbf{Outputs:} The proposed population of designs to evaluate $D_1,...,D_p$.
        \STATE Retrieve the population from the last generation $D_1^{'},...,D_p^{'}$ from $S$ (sorted by reward).
        \STATE Compute survival rate $x \leftarrow 0.6(1 - N_{cur}/N_{max})$
        \STATE Compute number of survivors from the last generation $p_{sur} \leftarrow \max{(2,\lceil px \rceil)}$
        \STATE \textbf{for} $i \leftarrow 1$ \TO $p_{sur}$ \textbf{do}
            \STATE \hskip1.0em $D_i \leftarrow D_i^{'}$
        \STATE \textbf{for} $i \leftarrow p_{sur}+1$ \TO $p$ \textbf{do}
            \STATE \hskip1.0em Randomly sample a parent $D_r^{'}$ from the survivors $D_1^{'},...,D_{p_{sur}}^{'}$
            \STATE \hskip1.0em Mutate $D_r^{'}$ to be $D_i$ with 10\% probability of changing each voxel
    \end{algorithmic}
\end{algorithm}

\subsection{Bayesian optimization (BO)}

BO tries to reduce the number of evaluations on expensive black-box functions by learning and utilizing a surrogate model. The surrogate model is learned, i.e. fitted by the history data, trying to map the inputs to their corresponding outputs, which mimics the real function evaluation. Instead of directly optimizing on the learned surrogate model, an acquisition function is constructed as the optimization objective in order to trade-off between exploitation and exploration of the surrogate model's prediction, which favors the uncertain region of the input space as well as the high-performing region. Next, the optimizer is applied on top of the acquisition function to search for the most promising input parameters to evaluate on the real function. Finally, the whole process repeats after the real evaluations are done and the results are added to the history data.

In our BO implementation, we use a Gaussian process as the surrogate model with a Matern 5/2 kernel \cite{10.5555/1162254}, Expected Improvement (EI) \cite{10.1007/3-540-07165-2_55} as the acquisition function, batch Thompson sampling \cite{russo2017tutorial} together with L-BFGS optimizer \cite{fletcher2013practical} to optimize the acquisition function in a batched manner. See Algorithm \ref{alg:bo} for more details in each generation of BO.


\begin{algorithm}
\caption{Bayesian optimization (per generation)}
    \label{alg:bo}
    \begin{algorithmic}
        \STATE \textbf{Inputs:} History data $S$, population size $p$.
        \STATE \textbf{Outputs:} The proposed population of designs to evaluate $D_1,...,D_p$.
        \STATE Fit a Gaussian process model $G$ that maps from designs $D$ to reward $r$ in dataset $S$
        \STATE Build expected improvement acquisition function $f$ based on prediction from $G$
        \STATE Generate initial population $D_1^{0},...,D_p^{0}$ by random sampling
        \STATE \textbf{for} $i \leftarrow 1$ \TO $p$ \textbf{do}
            \STATE \hskip1.0em Optimize design $D_i^{0}$ to be $D_i$ on acquisition function $f$ by L-BFGS
    \end{algorithmic}
\end{algorithm}

\subsection{CPPN-NEAT}

As described in Section \ref{sec:design_optim}, in this method, the robot design is parameterized by a Compositional Pattern Producing Network (CPPN) and NeuroEvolution of Augmenting Topologies (NEAT) algorithm is used to evolve the structure of CPPNs by working as a genetic algorithm with specific mutation, crossover, and selection operators defined on network structures. We use a standard implementation of both CPPN and NEAT components, so please refer to the original NEAT paper for the theoretical illustrations of the optimization process, and the hyperparameters of NEAT are presented in Appendix \ref{supp:hyperparam}.



\section{Hyperparameters}
\label{supp:hyperparam}
In this section we describe hyperparameters used for each algorithm in our work. We break down this section by first describing general hyperparameters used in our co-design experiments. Next, we specify the hyperparameters specific to each design optimization algorithm. Finally, we describe the hyperparameters used in proximal policy optimization (PPO), our control optimization algorithm.

\subsection{General hyperparameters}

\begin{table}[h!]
  \caption{Values of experiment hyperparameters}
  \label{tab:exphp}
  \centering
  \begin{tabular}{ll}
    \toprule
    \cmidrule(r){1-2}
    parameter name     & value \\
    \midrule
    population size & $25$\\
    robot shape & $(5\times 5)$, $(5\times 7)$ \\
    max evaluations & $250, 500, 750$\\
    train iters & $1000$\\
    \bottomrule
  \end{tabular}
\end{table}


In this section we describe some general hyperparameters used in all our co-design algorithms. Each of the co-design algorithms operates on a population of individuals, the size of which is specified by the parameter \textit{population size.} These algorithms also work with a grid-like design space whose size is specified by \textit{robot shape}. For most tasks the $(5\times 5)$ grid size is sufficient for the algorithms to find complex, interesting robots. The only exception is the \textit{Lifter} task where we increase the size of the design space to a $(5\times 7)$ grid, to accommodate a near-optimal robot we hand-designed. \textit{Max evaluations} specifies how many unique robots are trained in each algorithm. We use this metric to compare algorithms (instead of, for instance, a maximum number of generations) because the Genetic algorithm trains a different number of unique robots per generation compared to the Bayesian optimization and CPPN-NEAT algorithms. \textit{Max evaluations} varies between tasks as we use less evaluations to train algorithms whose performance converges faster. We do not train any algorithms more than $750$ evaluations. Finally, we train each robot for \textit{train iters} iterations using reinforcement learning (RL) in order to evaluate its performance on the task at hand. Note that the number of total RL steps will be the product of \textit{train iters}, \textit{num steps}, and \textit{num processes} (from Section \ref{hyper:ppo}) for a total of $512000$ steps.

\subsection{Design optimization hyperparameters}
In this section we specify hyperparameters relevant to our design optimization algorithms.

\subsubsection{Genetic algorithm (GA)}

\begin{table}[h!]
  \caption{Values of GA hyperparameters}
  \label{tab:gahp}
  \centering
  \begin{tabular}{ll}
    \toprule
    \cmidrule(r){1-2}
    parameter name     & value \\
    \midrule
        mutation rate & $10\%$\\
        survivor rate range & $[0.0, 0.6]$ \\
    \bottomrule
  \end{tabular}
\end{table}


The genetic algorithm only has two significant hyperparameters. The \textit{mutation rate} is important for constructing offspring robots of an existing survivor robot. The \textit{mutation rate} specifies the probability of mutating each voxel of the survivor robot's structure and the resulting structure after mutation becomes that of the offspring robot. The \textit{survivor rate range} specifies how the percent of robots that survive each generation of the algorithm changes over time. The survivor rate starts at the maximum value in the range and decreases linearly to the minimum value.

\subsubsection{Bayesian optimization (BO)}

\begin{table}[h!]
  \caption{Values of BO hyperparameters}
  \label{tab:bohp}
  \centering
  \begin{tabular}{ll}
    \toprule
    \cmidrule(r){1-2}
    parameter name     & value \\
    \midrule
        kernel variance $\sigma$ & $1.0$\\
        kernel length scale $l$ & $(1,...,1)\in \mathbb{R}^d$ \\
        optimizer max iterations & 100 \\
        optimizer restarts & 5\\
    \bottomrule
  \end{tabular}
\end{table}

We use the default implementation from the \texttt{GPyOpt} package and we do not change any specific hyperparameters. Values of the most important hyperparameters are listed in Table \ref{tab:bohp}. \textit{$\sigma$} and \textit{$l$} are the variance and the length scale of the Matern 5/2 kernel in the surrogate model, where $d$ is the dimension of the input (number of voxels). The \textit{optimizer max iterations} is the max number of iterations used to optimize the parameters of the surrogate model by the L-BFGS optimizer, and \textit{optimizer restarts} specifies the number of restarts in the optimization.

\subsubsection{CPPN-NEAT}

\begin{table}[h]
\caption{Values of CPPN-NEAT hyperparameters}
  \label{tab:cppnhp}
    \begin{subtable}[h]{0.45\textwidth}
        \centering
        \begin{tabular}{ll}
    \toprule
    \cmidrule(r){1-2}
    parameter name     & value \\
    \midrule
        pop size                & 50
        \\ 
        num inputs              & 3
        \\
        num hidden              & 1
        \\
        num outputs             & 5
        \\
        initial connection      & partial direct 0.5
        \\
        feed forward            & True
        \\
        compatibility threshold & 3.0
        \\
        compatibility disjoint coefficient    & 1.0
        \\
        compatibility weight coefficient      & 0.6
        \\
        conn add prob           & 0.2
        \\
        conn delete prob        & 0.2
        \\
        node add prob           & 0.2
        \\
        node delete prob        & 0.2
        \\
        activation options      & sigmoid
        \\
        activation mutate rate  & 0.0
        \\
        aggregation options     & sum
        \\
        aggregation mutate rate & 0.0
        \\
        bias init mean          & 0.0
        \\
        bias init stdev         & 1.0
        \\
        bias replace rate       & 0.1
        \\
    \bottomrule
  \end{tabular}
    \end{subtable}
    \hfill
    \begin{subtable}[h]{0.45\textwidth}
        \centering
        \begin{tabular}{ll}
    \toprule
    \cmidrule(r){1-2}
    parameter name     & value \\
    \midrule
        bias mutate rate        & 0.7
        \\
        bias mutate power       & 0.5
        \\
        bias max value          & 30.0
        \\
        bias min value          & -30.0
        \\
        response init mean      & 1.0
        \\
        response init stdev     & 0.0
        \\
        response replace rate   & 0.0
        \\
        response mutate rate    & 0.0
        \\
        response mutate power   & 0.0
        \\
        response max value      & 30.0
        \\
        response min value      & -30.0
        \\
        weight max value        & 30
        \\
        weight min value        & -30
        \\
        weight init mean        & 0.0
        \\
        weight init stdev       & 1.0
        \\
        weight mutate rate      & 0.8
        \\
        weight replace rate     & 0.1
        \\
        weight mutate power     & 0.5
        \\
        enabled default         & True
        \\
        enabled mutate rate     & 0.01
        \\
    \bottomrule
  \end{tabular}
    \end{subtable}
\end{table}

The hyperparameters of CPPN-NEAT are listed in Table \ref{tab:cppnhp}, whose interpretations can be found in the documentation of the neat-python package (\url{https://neat-python.readthedocs.io/en/latest/config_file.html}).

\subsection{Control optimization hyperparameters}
\label{hyper:ppo}

\begin{table}[h!]
  \caption{Values of PPO hyperparameters}
  \label{tab:ppohp}
  \centering
  \begin{tabular}{ll}
    \toprule
    \cmidrule(r){1-2}
    parameter name     & value \\
    \midrule
        use gae & True \\
        learning rate & $2.5 \cdot 10^{-4}$ \\
        use linear learning rate decay & True\\
        clip parameter & 0.1 \\
        value loss coefficient & 0.5 \\
        entropy coefficient & 0.01\\
        num steps & 128\\
        num processes & 4\\
        evaluation interval & 50\\
    \bottomrule
  \end{tabular}
\end{table}


In this section we specify hyperparameters relevant to our control optimization algorithm (PPO), which are listed in Table \ref{tab:ppohp}. The \textit{use gae} indicates whether we apply Generalized Advantage Estimation (GAE) \cite{schulman2015high} during PPO. The \textit{learning rate} is for the Adam optimizer \cite{kingma2017adam} optimizing the actor and critic networks and we \textit{use linear learning rate decay} throughout training. The \textit{clip parameter}, \textit{value loss coefficient}, \textit{entropy coefficient} are easily explained in any brief reference manual on PPO. The \textit{num steps} specifies the number of steps that each process samples in each iteration of PPO while the \textit{num processes} indicates the number of processes we use for parallel sampling. The \textit{evaluation interval} specifies the number of training iterations between evaluations. For all other parameters we use their default values. 

\section{Complete experiment results}
\label{supp:result}

In the main paper, we highlight ten tasks in our environment benchmark suite and compare their performance using all three design optimization algorithms.  Our full benchmark suite includes 32 tasks, described in detail in Section \ref{supp:benchmark}.  This sections details the evaluation results for all 32 tasks.

\begin{figure}[H]
    \centering
    \vspace{-1em}
    \includegraphics[width=0.8\textwidth]{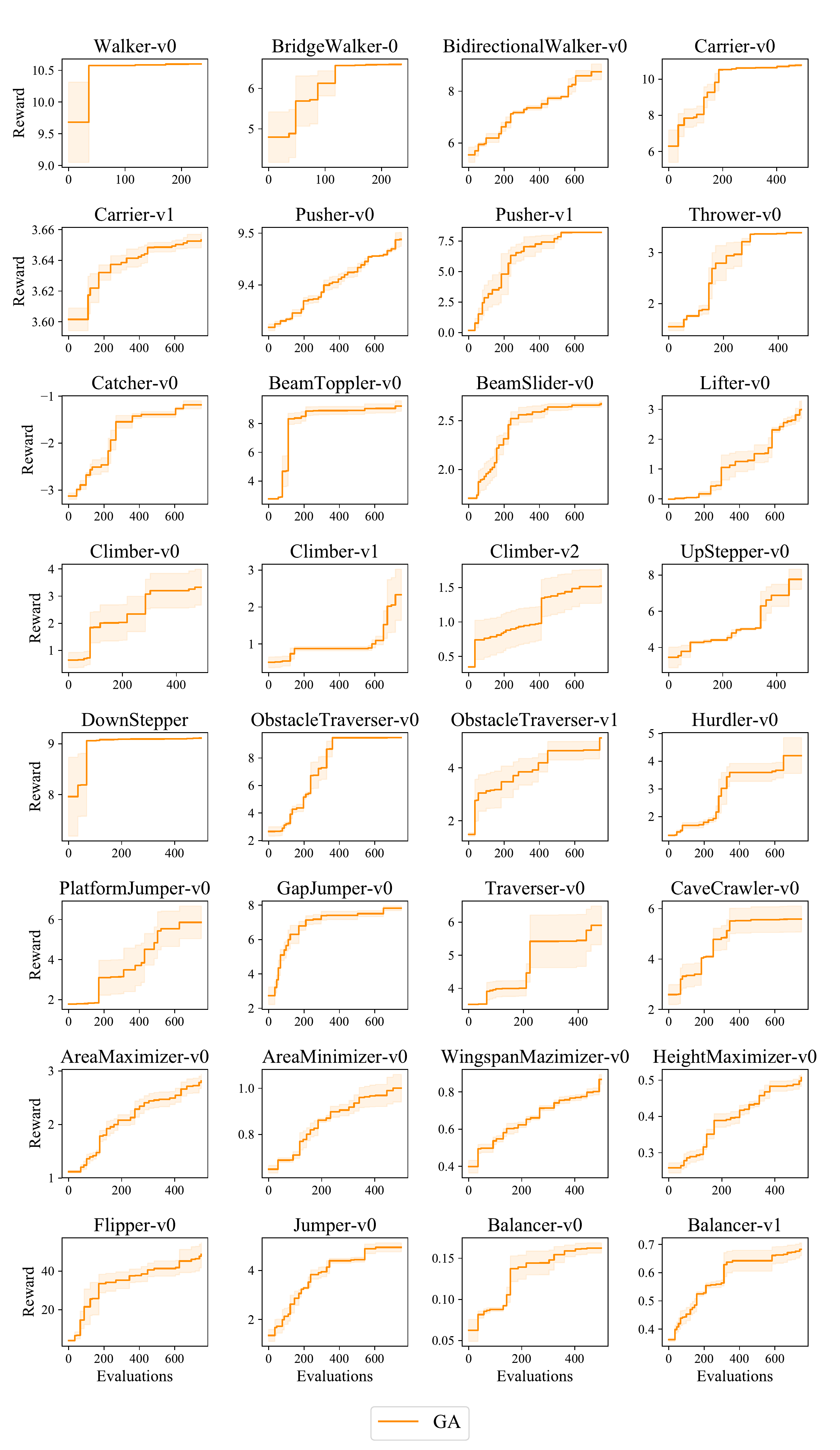}
    \caption{\textbf{Performance of GA baseline algorithm.} We plot the best performance of robots that the GA algorithm has evolved w.r.t. the number of evaluations on each task. All the curves are averaged over 3 different random seeds, and the variance is shown as a shaded region.}
    \label{fig:rewardcurves}
\end{figure}

We evaluated the complete benchmark suite using the baseline co-design algorithm with the GA for design optimization and PPO for the control optimization.  Figure \ref{fig:rewardcurves} shows the reward curves for the GA baseline algorithm on all 32 tasks.  In our main benchmark suite of ten tasks (see Section \ref{sec:experiment} of the main text), we found the GA baseline algorithm outperformed the other baselines algorithms the majority of the time.

\begin{figure}[H]
    \centering
    \vspace{-1em}
    \includegraphics[width=\textwidth]{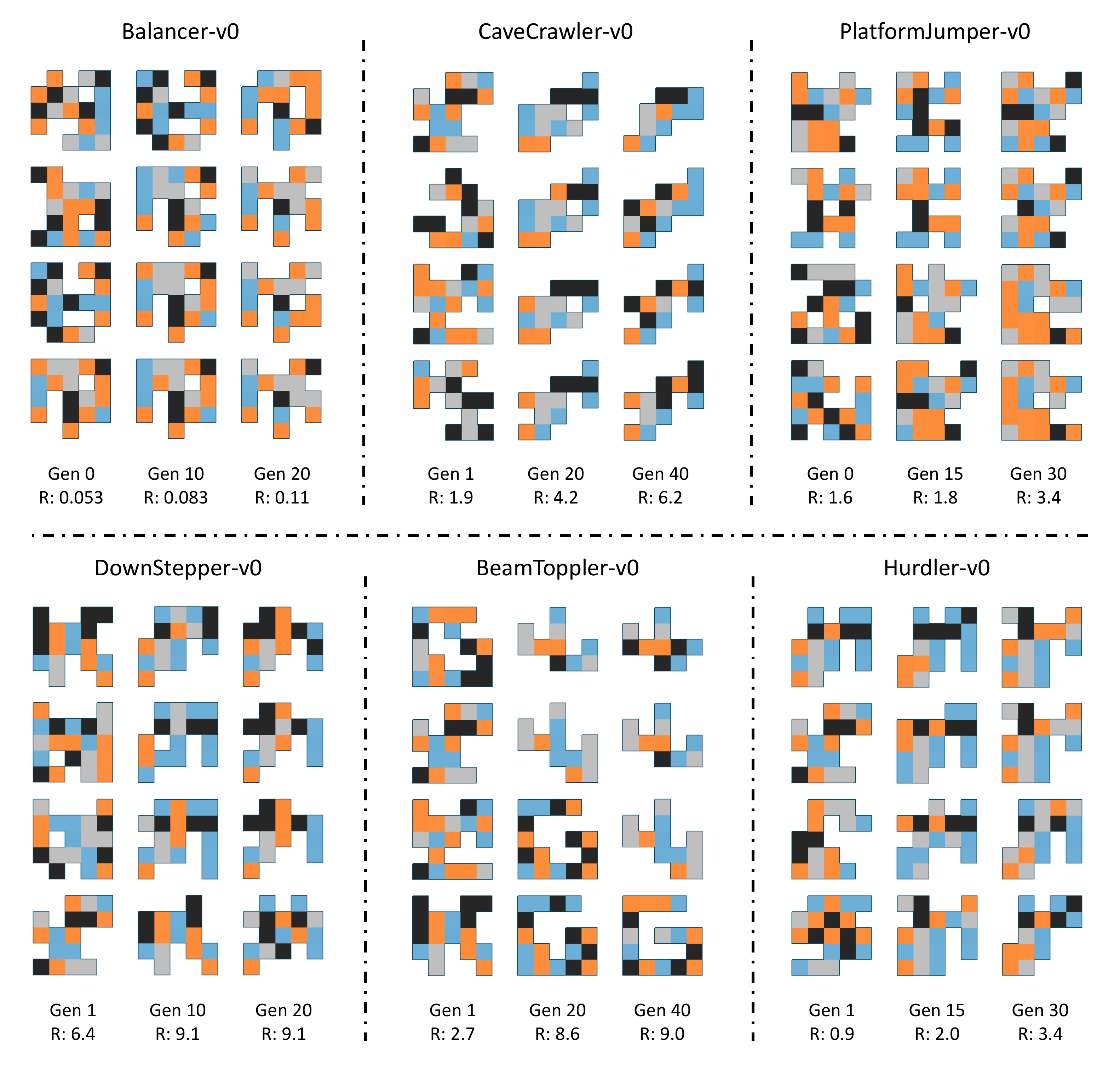}
    \caption{\textbf{Evolution of robot designs.} For each of the six selected tasks, we visualize the population in three different generations. Each column corresponds to one generation for which we show the four top performing robots along with their average reward.}
    \label{fig:evol}
\end{figure}

\begin{figure}[H]
    \centering
    \vspace{-1em}
    \includegraphics[width=\textwidth]{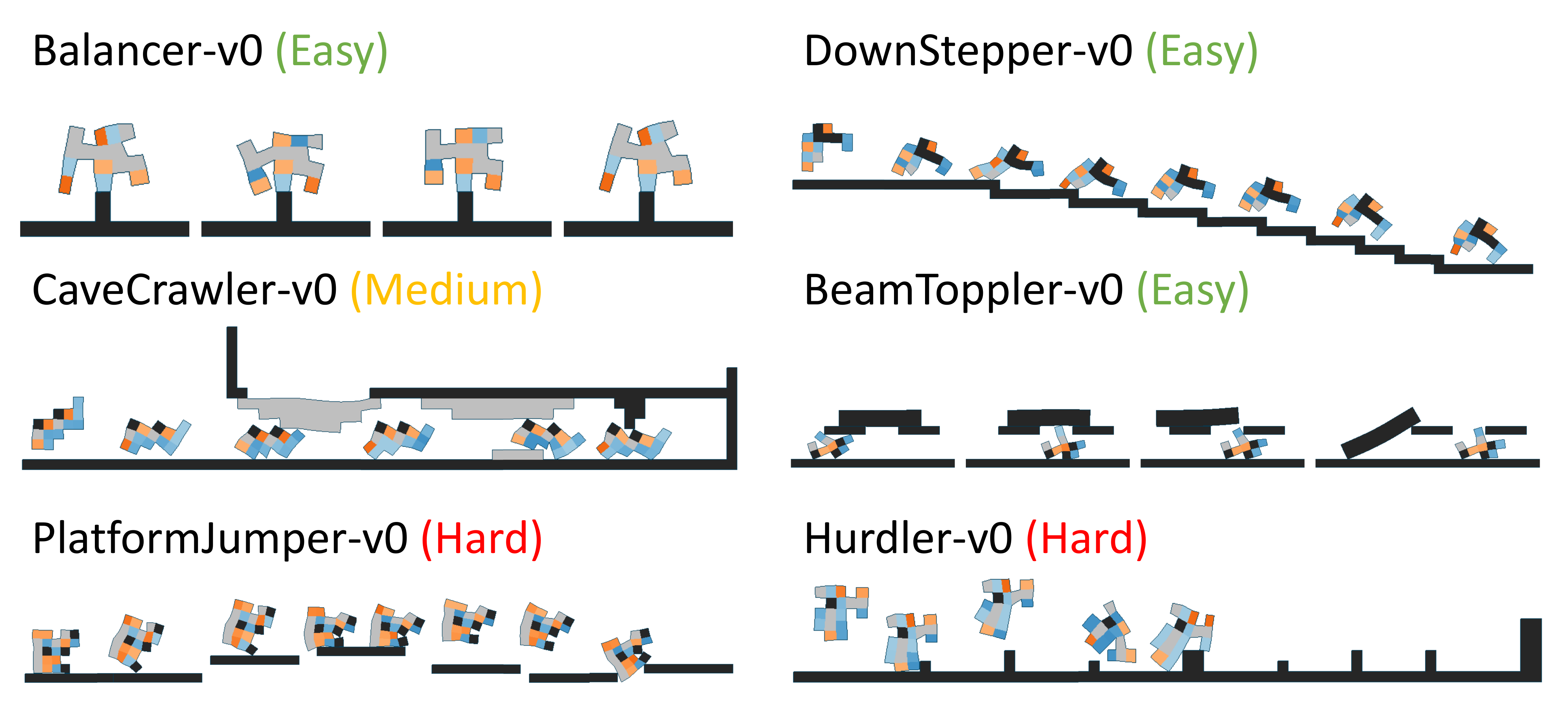}
    \caption{\textbf{Algorithm-optimized robots}. For each of the six selected tasks, we visualize a step-by-step sequence of a robot optimized by the algorithm.}
    \label{fig:results}
\end{figure}

We highlight the design and control optimization results of a subset of six tasks in Figures \ref{fig:evol} and \ref{fig:results}: Balancer-v0, CaveCrawler-v0, PlatformJumper-v0, DownStepper-v0, BeamToppler-v0, and Hurdler-v0.  In Figure \ref{fig:evol}, for each of the six tasks, we visualize the top four robots from three different generations. We also show the average reward these designs achieve.

Balancer-v0 requires the robot balance on top of a thin pole.  Figure \ref{fig:evol} reveals how the robot slowly evolves arm-like features which it actuates to maintain balance atop the pole -- much like how humans stretch out their arms to balance.

CaveCrawler-v0 requires the robot slither under and between a number of low-hanging obstacles.  By the last generation, the robots have converged on a small snake-like form which is short enough to clear rigid obstacles and is lined with horizontal actuators to allow slithering motion across the ground. 

PlatformJumper-v0 requires the robot jump between floating platforms at different heights.  Consequently, optimal designs contain many actuators oriented such that the robot can spring forward at will.

DownStepper-v0 requires the robot traverse down an uneven staircase.  Optimal robots evolve towards a bipedal form -- with a horizontal actuator on one foot and a vertical actuator on the other to promote seamless movement.

BeamToppler-v0 requires the robot knock over a beam resting on two pegs.  Interestingly, two optimal designs survive the co-design optimization.  Both designs evolve a hand-like mechanical arm to push the beam from underneath.  But one design (\#4 in generation 40, Figure \ref{fig:evol}) also evolves a hook-like gripper to push the beam off from above instead of from below.

In Figure \ref{fig:results} we show step-by-step sequences of the performance of six optimized designs.

A number of tasks perform very well.  Optimal robots in Balancer-v0 actuate their arm-like counterweights to maintain their position atop the thin pole.  Robots in DownStepper-v0 quickly run down the unevenly spaced stairs using their bipedal legs.  Optimal robots in BeamToppler-v0 repeatedly actuate their hand-like mechanism to nudge the beam of its pegs.

Some environments have few successful robots.  Most near-optimal designs produced in CaveCrawler-v0 are unable to clear the last rigid low-hanging obstacle in the cave.  Although some, like the robot shown in Figure \ref{fig:results} are able to clear all sections.

Some tasks are more complicated than others and the baseline algorithms fails to evolve a fully successful robot.  For example, the optimal robot in the PlatformJumper-v0 environment successfully lands on many platforms, but ultimately gets stuck in a gap between two platforms.  In Hurdler-v0, the robot is able to clear many tall thin vertical obstacle, but its hook-like jumping design ultimately fails when it gets caught on a wider vertical barrier.

A visualization of all robots for the 32 benchmark tasks is included in the Supplementary Materials.

\end{document}